\documentclass[preprint,12pt]{elsarticle}

\usepackage{amsmath}
\usepackage{amssymb}

\usepackage{graphicx}
\usepackage{float}
\usepackage{caption}
\usepackage{placeins}
\usepackage{algorithm}
\usepackage{algpseudocode}
\usepackage{array}
\usepackage{booktabs}
\usepackage{makecell}
\usepackage{ragged2e}
\usepackage{xltabular}
\usepackage{multirow}
\newcolumntype{L}[1]{>{\RaggedRight\arraybackslash}p{#1}}
\newcolumntype{Y}{>{\RaggedRight\arraybackslash}X}

\keepXColumns
\algrenewcommand\algorithmicrequire{\textbf{Input:}}
\algrenewcommand\algorithmicensure{\textbf{Output:}}

\journal{Computers in Industry}

\begin{document}

\begin{frontmatter}



\title{PD-SOVNet: A Physics-Driven Second-Order Vibration Operator Network for Estimating Wheel Polygonal Roughness from Axle-Box Vibrations}


\author[1]{Xiancheng Wang}
\ead{402196277@qq.com}

\author[1]{Lin Wang\corref{cor1}}
\ead{wanglin\_007@hitwh.edu.cn}

\author[1]{Rui Wang}
\ead{wangrui@hitwh.edu.cn}

\author[2]{Zhibo Zhang}
\ead{zhangzhibo@cqsf.com}

\author[1]{Minghang Zhao}
\ead{zhaomh@hit.edu.cn}

\author[1]{Xiaoheng Zhang}
\ead{3328212329@qq.com}

\author[1]{Zhongyue Tan}
\ead{13147268759@163.com}

\author[1]{Kaitai Mao}
\ead{3082994791@qq.com}

\affiliation[1]{organization={School of Ocean Engineering, Harbin Institute of Technology},
            addressline={West Wenhua Road}, 
            city={Weihai},
            postcode={264209}, 
            state={Shandong},
            country={China}}

\affiliation[2]{organization={Technical Center, Bogie Development Department, CRRC Qingdao Sifang Locomotive and Rolling Stock Co., Ltd},
            addressline={Jinhong East Road}, 
            city={Qingdao},
            postcode={266111}, 
            state={Shandong},
            country={China}}
\cortext[cor1]{Corresponding author}
\begin{abstract}

Quantitative estimation of wheel polygonal roughness from axle-box vibration signals is a challenging yet practically relevant problem for rail-vehicle condition monitoring. Existing studies have largely focused on detection, identification, or severity classification, while continuous regression of multi-order roughness spectra remains less explored, especially under real operational data and unseen-wheel conditions. To address this problem, this paper presents PD-SOVNet, a physics-guided gray-box framework that combines shared second-order vibration kernels, a $4\times4$ MIMO coupling module, an adaptive physical correction branch, and a Mamba-based temporal branch for estimating the 1st--40th-order wheel roughness spectrum from axle-box vibrations. The proposed design embeds modal-response priors into the model while retaining data-driven flexibility for sample-dependent correction and residual temporal dynamics. Experiments on three real-world datasets, including operational data and real fault data, show that the proposed method provides competitive prediction accuracy and relatively stable cross-wheel performance under the current data protocol, with its most noticeable advantage observed on the more challenging Dataset III. Noise injection experiments further indicate that the Mamba temporal branch helps mitigate performance degradation under perturbed inputs. These results suggest that structured physical priors can be beneficial for stabilizing roughness regression in practical rail-vehicle monitoring scenarios, although further validation under broader operating conditions and stricter comparison protocols is still needed.

\end{abstract}


\begin{highlights}
\item A physics-guided gray-box framework is proposed for regressing multi-order wheel polygonal roughness from axle-box vibration signals.
\item Shared second-order vibration kernels, a $4\times4$ MIMO coupling module, adaptive physical correction, and a Mamba temporal branch are integrated within a unified regression architecture.
\item On three real-world datasets and an unseen-wheel evaluation protocol, the method shows competitive accuracy and relatively stable cross-wheel performance, especially on the more challenging dataset.
\end{highlights}


\end{frontmatter}


\section{Introduction}

Wheel polygonization is a typical wheel--rail issue that occurs during the service of high-speed railway and urban rail transit vehicles. It is mainly manifested as periodic geometric irregularities along the rolling circumference of the wheel. Such irregularities can induce sustained periodic dynamic responses during wheel--rail contact, thereby intensifying wheel--rail impacts, aggravating axle-box and bogie-frame vibrations, and potentially accelerating fatigue damage and performance degradation of related components \cite{nielsen2000out,tao2020polygonisation}. With the increase in train operating speed and the growing demand for intelligent operation and maintenance, the use of measurable onboard vibration signals for wheel condition monitoring and quantitative evaluation has gradually become an important topic in rail vehicle health monitoring \cite{alemi2017condition}.

\begin{figure}[H]
    \centering
    \includegraphics[width=0.9\linewidth]{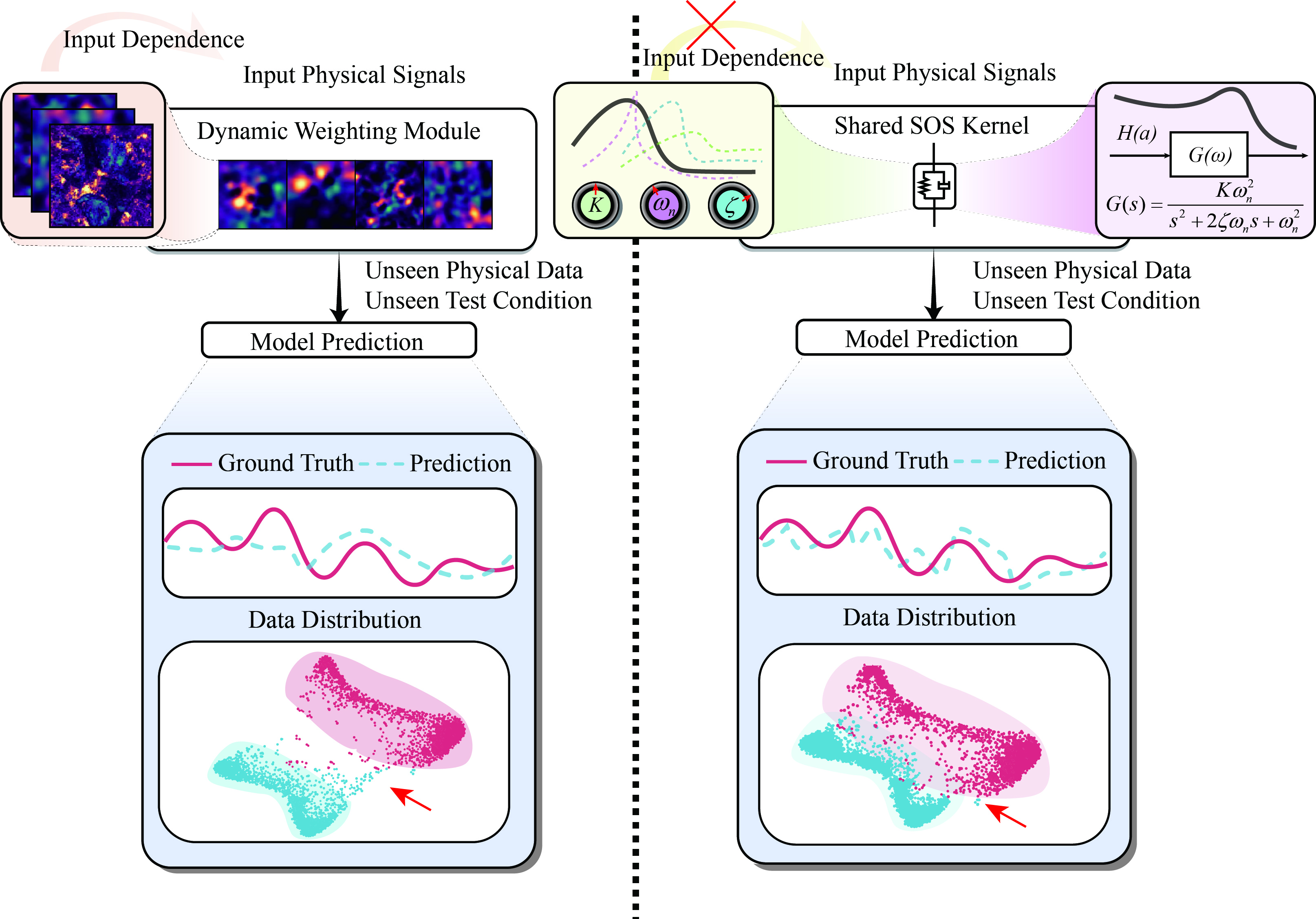}
    \caption{Schematic comparison of the regression behavior of a purely data-driven network and a physics-driven second-order vibration operator network under the current unseen-wheel evaluation setting. The method on the left mainly relies on input-dependent dynamic weighting mechanisms to learn the mapping relationship and may therefore depend more heavily on the training distribution when the response--label mapping varies across wheel groups. The method on the right explicitly embeds physical priors, such as gain, natural frequency, and damping, into the model through shared second-order vibration operator kernels, with the aim of making prediction behavior more stable under the current data protocol.}
    \label{fig:intro_compare}
\end{figure}

Considerable progress has been made in studies of wheel polygonization, including its formation mechanism, monitoring methods, and engineering applications. Because axle-box vibration signals can more directly reflect the dynamic characteristics of the wheel--rail contact process, they have become a commonly used source of information for wheel polygon monitoring. Existing methods are mostly based on time--frequency analysis, angle-domain synchronous averaging, parametric spectral estimation, and non-stationary signal decomposition, through which polygon-related response features are extracted from vibration signals to achieve order identification, anomaly detection, or condition classification \cite{song2020railway,sun2021board,chen2022quantitative}. These methods have shown good engineering practicality in detection and identification tasks, but overall they still mainly focus on condition recognition.

In practical operation and maintenance scenarios, however, simply determining whether a wheel exhibits polygonization is often insufficient for refined maintenance decision-making. For wheelset health management, roughness evolution analysis, and maintenance-level assessment, it is more valuable to further estimate wheel surface roughness from vibration responses and to obtain the roughness distribution over different orders. In other words, the task is closer to a continuous regression problem from axle-box vibration signals to the multi-order wheel roughness spectrum. In recent years, with the increasing use of deep learning in complex nonlinear mapping problems, some studies have attempted to introduce convolutional neural networks or hybrid deep models into out-of-round wheel monitoring in order to enhance the automatic extraction and representation of complex patterns \cite{ye2022oornet,dong2024quantitative,sun2025board}. Nevertheless, existing work still focuses largely on fault diagnosis, anomaly detection, or severity classification, and there remains room for further development in the continuous regression and robust estimation of wheel polygon roughness values.

The difficulty of this problem lies in the fact that the relationship between wheel roughness and axle-box vibration response is not a simple direct mapping. The propagation of roughness-induced excitation in the wheel--rail system is jointly affected by wheel--rail contact conditions, structural modal responses, vibration transmission paths, and multi-wheel coupling effects. In addition, variations in dynamic boundary conditions across different wheels, bogies, and operating conditions can further shift this mapping relationship \cite{li2025model}. As a result, although traditional signal processing methods are suitable for polygon detection and order localization, their ability to represent continuous roughness regression is relatively limited. Pure deep learning methods, by contrast, possess strong fitting capability, but they often rely on black-box networks to implicitly learn the transmission paths, resonance responses, and the relationship between roughness values and vibration amplitudes. Such methods can achieve good performance when the training data provide sufficient coverage, but their stability and physical interpretability under cross-wheel and unseen-wheel scenarios still leave room for improvement.

In recent years, physics-driven machine learning and physics-informed neural networks have provided new ideas for modeling and inference in complex engineering systems \cite{karniadakis2021physics,cuomo2022scientific,raissi2019physics}. Existing studies have shown that explicitly embedding physical laws, prior structures, or operator constraints into neural networks may help improve model robustness, interpretability, and stability while retaining the flexibility of data-driven methods \cite{lu2021learning,li2020fourier,wang2021learning,li2024physics}. Although such studies have mainly focused on partial differential equations and scientific computing, their core ideas are also relevant to wheel polygon roughness estimation. For a wheel--rail system with clear vibration transmission mechanisms and order-response characteristics, explicitly incorporating second-order vibration response priors into the network architecture may help construct a more stable roughness regression model with improved physical consistency.

Motivated by this, this paper proposes PD-SOVNet (\textit{Physics-Driven Second-Order Vibration Operator Network}) for estimating wheel roughness values at 40 orders from axle-box vibration data. Unlike purely data-driven methods, the proposed model embeds shared second-order vibration kernels into the network architecture to characterize vibration amplification, attenuation, and resonance-response characteristics around different orders. A $4\times4$ MIMO channel coupling module is further introduced to model cross-channel transmission relationships, and a Mamba temporal branch is incorporated to supplement residual temporal dynamics not fully covered by the physical branch. Rather than claiming a complete physical reconstruction of the wheel--rail system, the proposed design aims to investigate whether introducing structured vibration priors can improve prediction stability and interpretability under the current unseen-wheel evaluation setting.

As illustrated in Fig.~\ref{fig:intro_compare}, purely data-driven models may become more dependent on the training distribution when the mapping between vibration responses and roughness spectra varies across wheel groups. The motivation of PD-SOVNet is therefore not to claim universal superiority over black-box models, but to examine whether explicitly introducing shared vibration-response priors can make the regression behavior more stable under the current data protocol and unseen-wheel evaluation setting.

Table~\ref{tab:lit_supp} summarizes representative studies closely related to this work.

\begingroup
\footnotesize
\setlength{\tabcolsep}{5pt}
\renewcommand{\arraystretch}{1.18}

\begin{xltabular}{\textwidth}{L{2.45cm} Y Y}
\caption{Representative studies related to wheel polygonization and physics-driven modeling}
\label{tab:lit_supp}\\

\toprule
References & Method / Research focus & Relevance to this study \\
\midrule
\endfirsthead

\caption[]{Representative studies related to wheel polygonization and physics-driven modeling (continued)}\\
\toprule
References & Method / Research focus & Relevance to this study \\
\midrule
\endhead

\bottomrule
\endfoot

\multicolumn{3}{@{}l}{\textbf{A. Key studies on wheel polygonization}}\\
\midrule

\parbox[t]{\linewidth}{Iwnicki et al.\\(2023)\\\cite{iwnicki2023out}}
& Reviews out-of-round wheels and wheel polygonization, covering dynamic mechanisms, detection methods, and engineering applications.
& Provides support for understanding recent developments in the field and can be used to strengthen the research background and engineering significance of this study.\\

\parbox[t]{\linewidth}{Wang et al.\\(2023)\\\cite{wang2023dynamic}}
& Proposes a dynamic detection method for wheel polygonization based on parametric power spectral estimation, improving the identification of weak features.
& Represents a typical traditional frequency-domain enhancement route and highlights the advantages of classical signal-processing methods in detection tasks.\\

\parbox[t]{\linewidth}{Guedes et al.\\(2023)\\\cite{guedes2023detection}}
& Combines wayside monitoring with artificial intelligence methods for wheel polygonization identification.
& Forms a useful comparison with the onboard axle-box-vibration-based scheme adopted in this work and illustrates differences among monitoring deployment modes.\\

\parbox[t]{\linewidth}{Magalh\~{a}es et al.\\(2024)\\\cite{magalhaes2024strategy}}
& Uses sparse autoencoders to identify out-of-round wheel damage while reducing dependence on labeled data.
& Serves as a relevant reference for weakly supervised or unsupervised approaches and shows the potential of data-driven methods under limited-label conditions.\\

\parbox[t]{\linewidth}{Jiang et al.\\(2025)\\\cite{jiang2025structure}}
& Proposes a structure-assisted generalized diagnosis network with emphasis on cross-scenario diagnostic capability.
& Is highly relevant to the cross-wheel generalization objective of this study and supports the focus on generalization performance.\\

\addlinespace[2pt]
\multicolumn{3}{@{}l}{\textbf{B. Key studies on physics-driven learning and neural operators}}\\
\midrule

\parbox[t]{\linewidth}{Wang et al.\\(2023)\\\cite{wang2023expert}}
& Summarizes training experience, common issues, and practical recommendations for PINNs.
& Provides methodological guidance for training physics-driven models and highlights the importance of reasonable training strategies.\\

\parbox[t]{\linewidth}{Kontolati et al.\\(2024)\\\cite{kontolati2024learning}}
& Learns nonlinear operators in latent space for efficient prediction of complex physical systems.
& Offers inspiration for low-dimensional representation of complex dynamic responses and helps explain efficient representation issues in physical systems.\\

\parbox[t]{\linewidth}{Yin et al.\\(2024)\\\cite{yin2024scalable}}
& Studies a scalable learning framework for geometry-dependent PDE solution operators.
& Suggests that operator learning has strong generalization potential across structure-varying and geometry-varying tasks.\\

\parbox[t]{\linewidth}{Wang et al.\\(2025)\\\cite{wang2025kolmogorov}}
& Introduces KAN into a physics-informed learning framework to enhance model expressiveness.
& Indicates that physics-driven networks can be combined with stronger function-approximation units, providing ideas for future model extension.\\

\parbox[t]{\linewidth}{Jiao et al.\\(2021)\\\cite{jiao2021one}}
& Investigates the one-shot learning capability of PDE solution operators and emphasizes the importance of prior structure.
& Further supports the view that appropriate priors can improve data efficiency and generalization, which is consistent with the modeling motivation of this study.\\

\end{xltabular}
\endgroup

The main contributions of this paper are as follows:
\begin{itemize}
    \item For the quantitative roughness-regression task in wheel polygon monitoring, a physics-guided gray-box framework is proposed to infer multi-order wheel roughness from axle-box vibration signals under an unseen-wheel evaluation protocol.
    \item A network structure integrating shared second-order vibration kernels, a $4\times4$ MIMO coupling module, adaptive physical correction, and a Mamba temporal branch is designed to encode modal priors, cross-channel interactions, and residual temporal dynamics within a unified regression model.
    \item Experiments on three real-world datasets suggest that this gray-box design provides competitive accuracy and relatively stable cross-wheel performance under the current data protocol, with the most noticeable benefit appearing on the more challenging dataset.
\end{itemize}
\section{Method}

\subsection{Data preprocessing}\label{数据预处理}

\begin{figure}[H]
    \centering
    \includegraphics[width=0.7\textwidth]{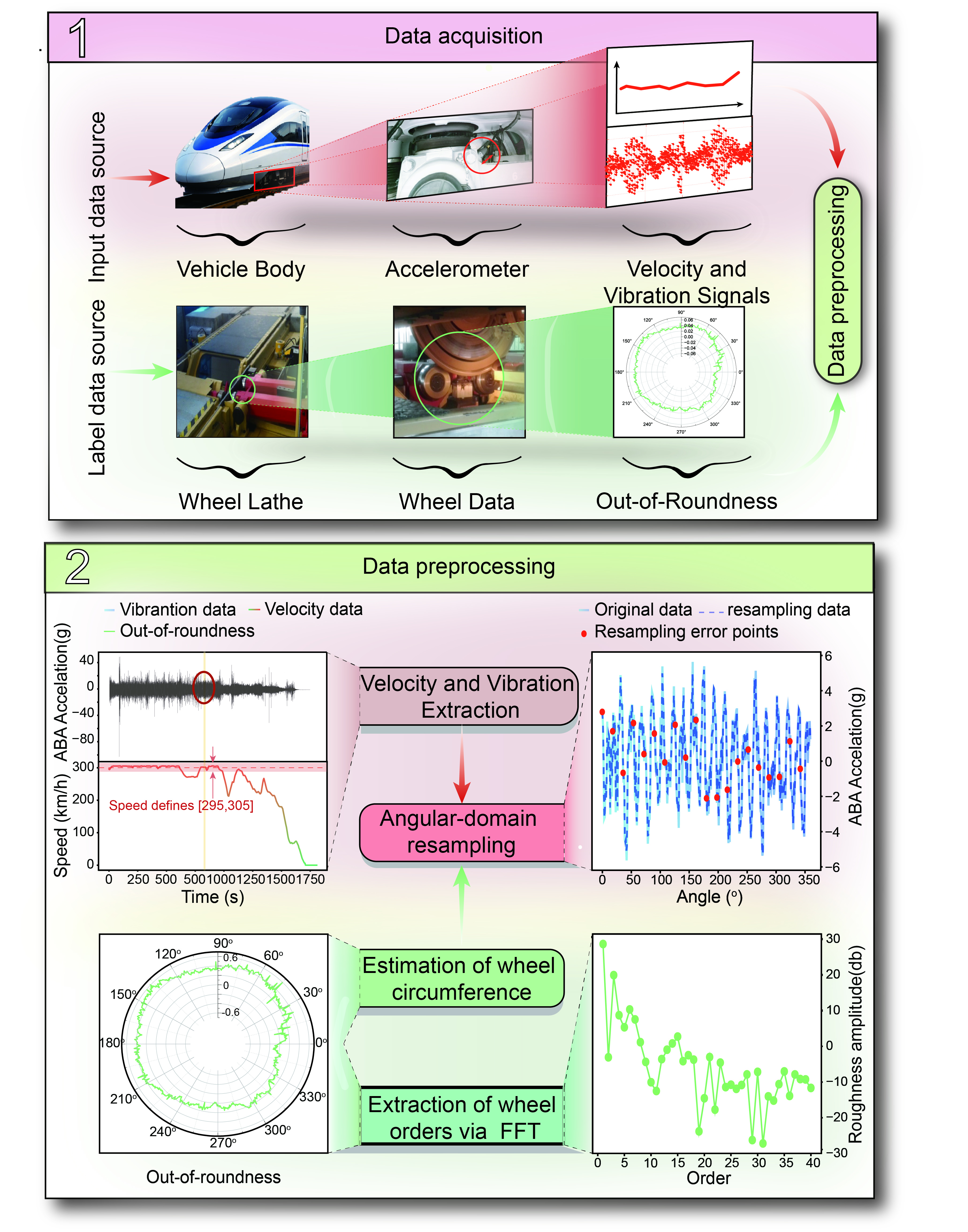}
    \caption{Schematic illustration of the data acquisition and preprocessing pipeline. The upper part shows the sources of the input data, namely axle-box vibration and speed signals collected during vehicle operation; the lower part shows the source of the label data, namely wheel out-of-roundness profiles measured by an underfloor wheel lathe. After speed filtering, angle-domain resampling, and order extraction, the input samples and the 1--40-order roughness labels for supervised regression are finally constructed.}
    \label{fig:data_pipeline}
\end{figure}

The data used in this study include both input signals and supervision labels, as shown in Fig.~\ref{fig:data_pipeline}. The input data are obtained from axle-box vibration and speed records during vehicle operation. Among them, the axle-box vibration signals are collected by accelerometers installed at the axle-box positions, with a sampling frequency of 10~kHz. The speed signal is used for subsequent operating-condition filtering and sample alignment, with a sampling frequency of 1~Hz. The label data are obtained from wheel measurements on an underfloor wheel lathe. By performing frequency-domain analysis on the wheel out-of-roundness profile along the circumference, roughness labels related to wheel polygonization are constructed. In this way, a supervised regression data pair between operational vibration responses and wheel roughness spectra is established.

The wheel roughness labels were obtained from wheel-profile measurements and matched to the corresponding wheel groups according to the current data-management protocol used in this study. Because operational vibration records and wheel-profile measurements are difficult to acquire in a fully synchronous manner in practical maintenance workflows, a certain level of input--label uncertainty may remain. Therefore, the reported regression results should be interpreted together with this practical limitation, which is also one of the reasons why the paper emphasizes stability under the current data protocol rather than making broader claims beyond the available data coverage.

\begin{figure}[t]
    \centering
    \includegraphics[width=1\linewidth]{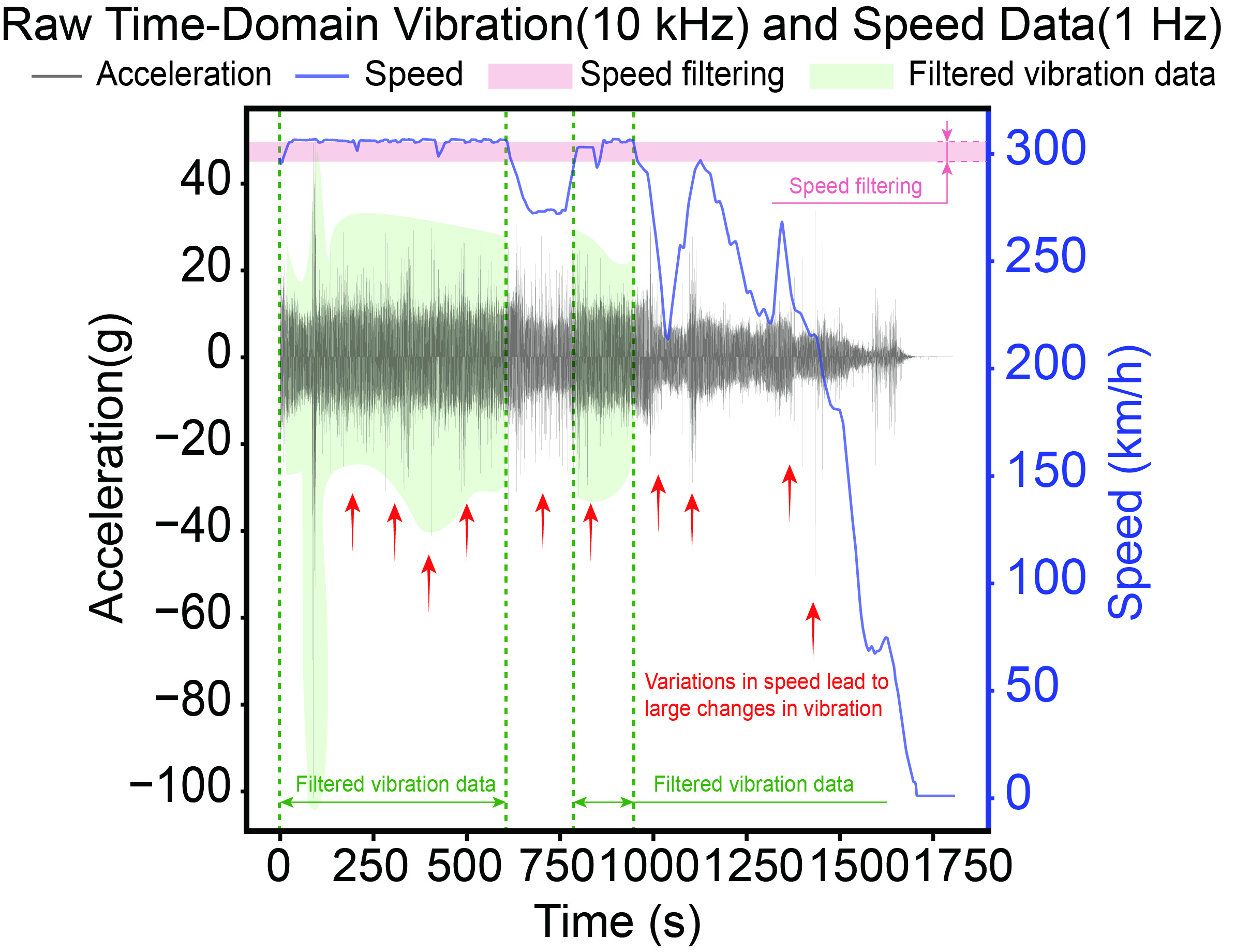}
    \caption{Illustration of the raw time-domain vibration and speed data. The left axis represents vibration acceleration, and the right axis represents operating speed. It can be seen that although the speed is relatively stable in some high-speed intervals, the vibration response still changes significantly. Therefore, speed filtering is first required before subsequent modeling.}
    \label{fig:speed_filter}
\end{figure}

\begin{figure}[htbp]
    \centering
    \includegraphics[width=0.7\linewidth]{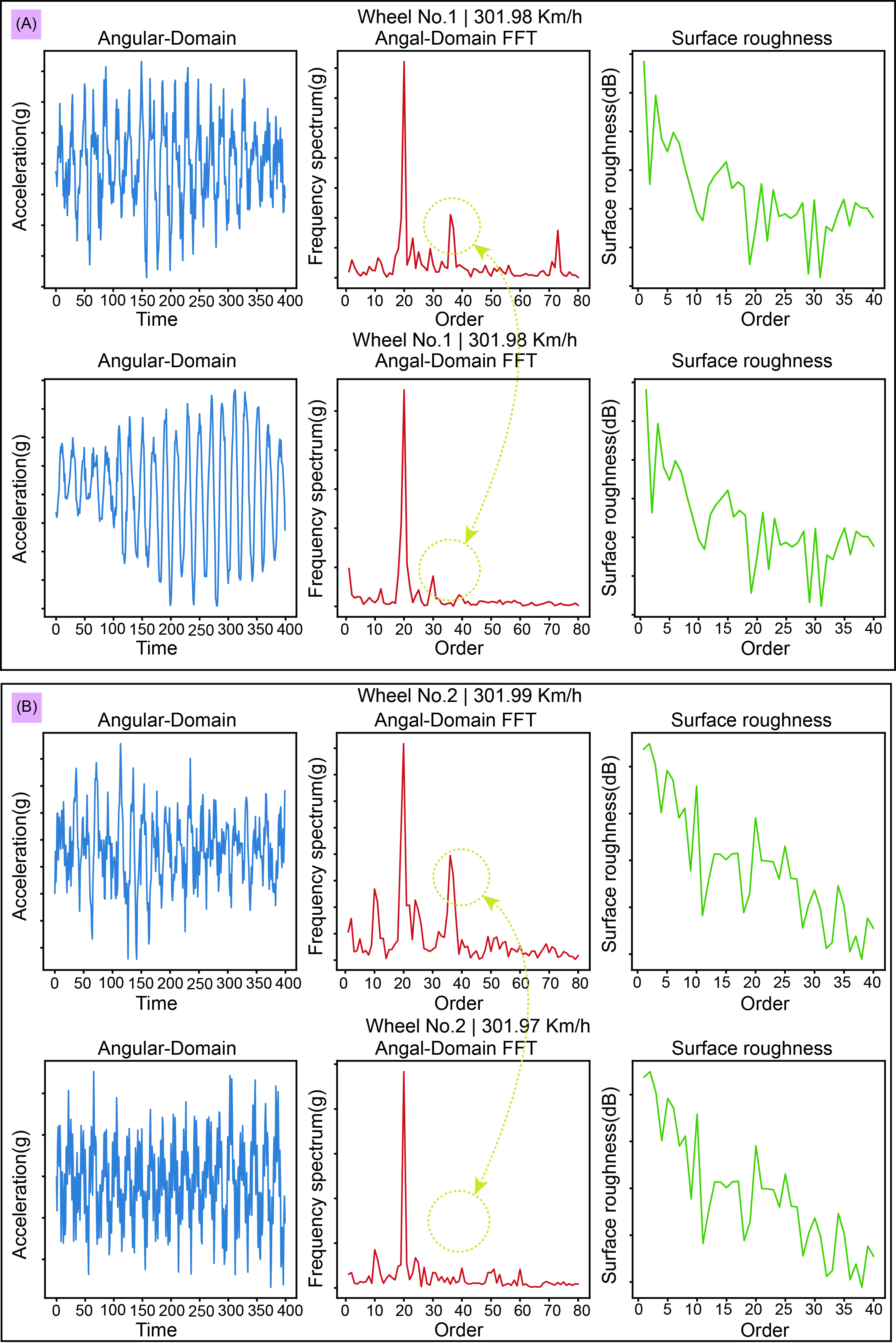}
    \caption{Examples of angle-domain resampled samples. Panel A shows the angle-domain vibration, the corresponding order spectrum, and the roughness labels of Wheel No.1 under similar speed conditions; Panel B shows the angle-domain vibration, the corresponding order spectrum, and the roughness labels of Wheel No.2 under similar speed conditions. The yellow dashed boxes indicate regions with relatively obvious differences in the order domain, showing that even under similar speed conditions, the vibration patterns and order responses may still differ significantly across wheels. Note: the vertical-axis values are masked due to data-sensitivity requirements from the data provider.}
    \label{fig:data_example}
\end{figure}

Since axle-box vibration is sensitive to operating speed, the signal distributions under different speed conditions may differ significantly. Therefore, the raw data are first filtered according to operating conditions. As shown in Fig.~\ref{fig:speed_filter}, although the train speed is relatively stable in some high-speed intervals, it varies considerably in other intervals, and even within stable-speed windows, the vibration response still shows strong fluctuations. To reduce the additional disturbance introduced by speed variation to the vibration distribution, only valid samples under the target high-speed condition are retained, and the speed range is limited to 295--305~km/h. The specific differences can be seen in Fig.~\ref{fig:data_example}. This treatment allows the subsequent model to focus more on response differences caused by wheel conditions themselves, rather than distribution shifts caused by speed variation.

After speed filtering, the retained vibration segments are further transformed into the angle domain to characterize response features directly related to the wheel rotation cycle. Specifically, the single-revolution vibration segment is first extracted according to the speed information. Then, together with the wheel circumference information obtained from a complete revolution measured on the wheel lathe, each revolution is resampled into the angle domain, as shown in Step 2 of Fig.~\ref{fig:data_pipeline}, so that each single-revolution vibration signal is represented as a fixed-length sequence. In this study, each revolution sample is resampled into 400 angular points, as shown in Step 2 of Fig.~\ref{fig:data_pipeline}, thereby yielding a unified angle-domain input representation. Meanwhile, fast Fourier transform is applied to the wheel out-of-roundness profile measured by the wheel lathe, and the roughness values of Orders 1--40 are extracted as supervision labels.

As shown in Fig.~\ref{fig:data_example}, after angle-domain resampling, different samples can all be represented as unified single-revolution angle-domain vibration sequences of the same length. Furthermore, frequency-domain analysis shows that different wheels may exhibit significant differences around certain orders. It is worth noting that even under similar speed conditions, the angle-domain waveforms, order spectra, and roughness distributions of different wheels still exhibit considerable complexity. This also indicates that the task is not a simple speed-driven mapping, but instead requires simultaneous modeling of the structural response of the wheel--rail system and cross-wheel differences.

In the specific implementation, the four wheels of one bogie are taken as the basic organizational unit, corresponding to the four channels of front-left, front-right, rear-left, and rear-right, respectively. For each sample, the raw angle-domain vibration input is denoted by $\mathbf{x}\in\mathbb{R}^{400\times4}$, where 400 denotes the number of angular points after single-revolution resampling and 4 denotes the four wheel channels. Furthermore, the corresponding order-domain representation is denoted by $\mathbf{x}_{\mathrm{order}}\in\mathbb{R}^{40\times4}$, representing the order-response features of the input signal at the 40 target orders. The speed information is denoted by $\mathbf{v}\in\mathbb{R}^{1\times4}$, and the supervision label is denoted by $\mathbf{y}\in\mathbb{R}^{40\times4}$, representing the roughness labels of the four wheels at Orders 1--40. Therefore, the data preprocessing pipeline in this study can be summarized as
\[
\left(\mathbf{x}_{\mathrm{raw}},\mathbf{v}_{\mathrm{raw}},\mathbf{r}_{\mathrm{wheel}}\right)
\rightarrow
\left(\mathbf{x}_{\mathrm{seg}},\mathbf{v}_{\mathrm{seg}}\right)
\rightarrow
\mathbf{x}
\rightarrow
\mathbf{x}_{\mathrm{order}};
\mathbf{r}_{\mathrm{wheel}}
\rightarrow
\mathbf{y},
\]
where $\mathbf{x}_{\mathrm{raw}}$ and $\mathbf{v}_{\mathrm{raw}}$ denote the raw vibration and speed data, respectively, and $\mathbf{r}_{\mathrm{wheel}}$ denotes the wheel out-of-roundness profile; $\mathbf{x}_{\mathrm{seg}}$ and $\mathbf{v}_{\mathrm{seg}}$ denote the valid segments after speed filtering; $\mathbf{x}$ denotes the single-revolution input after angle-domain resampling; $\mathbf{x}_{\mathrm{order}}$ denotes its corresponding order-domain representation; and $\mathbf{y}$ denotes the 1--40-order roughness labels obtained from the frequency-domain analysis of the wheel profile. Through the above processing, the raw operational data are uniformly mapped into an input--label space suitable for the subsequent physics-driven modeling.
\subsection{Order Frontend and Order Mixing Modules}

\begin{figure*}[!htbp]
    \vspace{-80pt}
    \centering
    \includegraphics[width=1\textwidth]{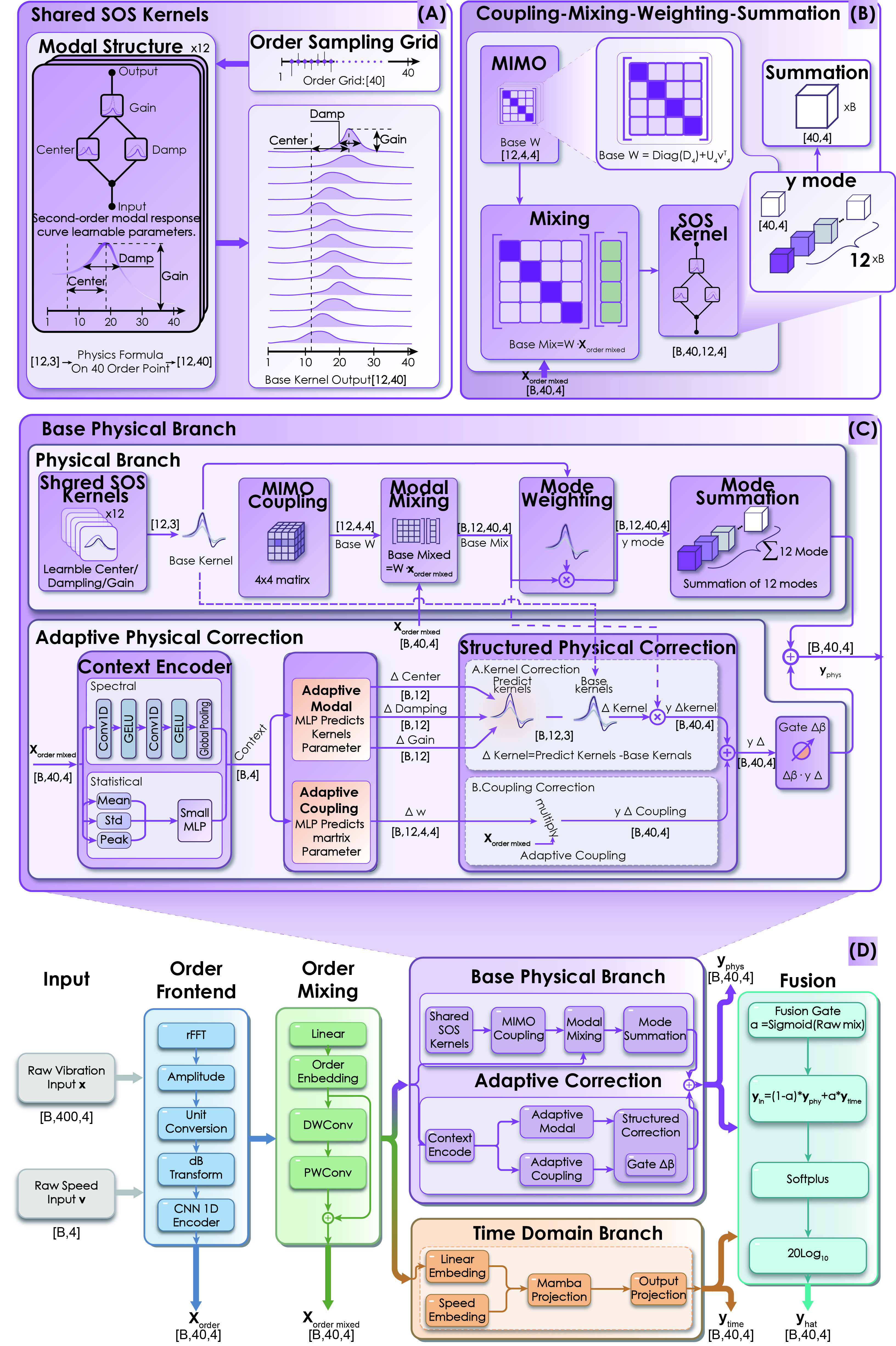}
    \caption{Overall architecture of PD-SOVNet. (A) Shared second-order vibration kernels. (B) MIMO coupling with modal mixing, weighting, and summation. (C) Physics-driven branch with adaptive physical correction. (D) Overall workflow of the proposed framework.}
    \label{fig:model_structure}
\end{figure*}

As shown in Part D of Fig.~\ref{fig:model_structure}, before entering the physics-driven branch, the model first performs order representation construction and local order mixing on the raw angle-domain vibration signal. The main purpose of this part is to map the input signal into an order-domain representation space that is more suitable for subsequent physics-based modeling, while enhancing the local correlations among neighboring orders. Since these two modules are not the core innovations of this work, only their overall logic is presented here without going into excessive implementation details.
\clearpage
\clearpage
Given an input sample $\mathbf{x}\in\mathbb{R}^{400\times4}$ and $\mathbf{v}\in\mathbb{R}^{4}$, where $\mathbf{x}$ denotes the single-revolution angle-domain vibration sequence with four channels and $\mathbf{v}$ denotes the corresponding speed information, the role of the Order Frontend is to transform the raw input into representation features at 40 target orders, namely
\[
\mathbf{x}_{\mathrm{order}}=\mathcal{F}_{\mathrm{front}}(\mathbf{x},\mathbf{v}),
\qquad
\mathbf{x}_{\mathrm{order}}\in\mathbb{R}^{40\times4}.
\]
As can be seen from Fig.~\ref{fig:model_structure}, this process mainly includes frequency-domain transformation, amplitude extraction, unit mapping, dB transformation, and lightweight one-dimensional convolutional encoding. Its core purpose is to convert the raw angle-domain vibration signal into a unified order-domain input representation, thereby providing the input basis for the subsequent physical response modeling using shared second-order vibration kernels.

After obtaining the initial order features, the model further introduces an Order Mixing module to enhance the local correlations in the order domain. Considering that neighboring orders often exhibit strong local relationships in wheel--rail vibration responses, this study adopts lightweight local convolutional mixing instead of introducing a more complex global attention structure. Specifically, $\mathbf{x}_{\mathrm{order}}$ is first projected into a higher-dimensional latent space and added with learnable order embeddings. Local convolution is then used to extract local patterns among neighboring orders, after which the features are projected back to the original channel dimension. The mixed order representation is finally obtained in a residual manner:
\[
\mathbf{h}=\mathrm{Linear}(\mathbf{x}_{\mathrm{order}})+\mathbf{E}_{\mathrm{order}},
\]
\[
\Delta\mathbf{x}=\mathrm{Proj}\!\left(\mathrm{LocalConv}(\mathbf{h})\right),
\]
\[
\mathbf{x}_{\mathrm{order}}^{\mathrm{mix}}
=
\mathbf{x}_{\mathrm{order}}
+\lambda\,\Delta\mathbf{x},
\qquad
\mathbf{x}_{\mathrm{order}}^{\mathrm{mix}}\in\mathbb{R}^{40\times4},
\]
where $\mathbf{E}_{\mathrm{order}}$ denotes the learnable order embedding and $\lambda$ is a learnable scaling coefficient. Through this design, Order Mixing preserves the original order information while smoothly modeling the local structure among neighboring orders, thereby providing a more stable input representation for subsequent modal-kernel weighting and channel coupling.

The above process can be summarized in Algorithm~\ref{alg:order_mixing}.

\begin{algorithm}[t]
\caption{Order Frontend and Order Mixing}
\label{alg:order_mixing}
\begin{algorithmic}[1]
\Require angle-domain vibration $\mathbf{x}$, speed information $\mathbf{v}$
\Ensure mixed order representation $\mathbf{x}_{\mathrm{order}}^{\mathrm{mix}}$
\State Use the Order Frontend to map $(\mathbf{x},\mathbf{v})$ into order-domain features $\mathbf{x}_{\mathrm{order}}$
\State Apply linear projection to $\mathbf{x}_{\mathrm{order}}$ and add order embeddings
\State Use local convolution to extract local patterns among neighboring orders
\State Project the convolution output back to the original channel dimension
\State Obtain $\mathbf{x}_{\mathrm{order}}^{\mathrm{mix}}$ through a residual connection
\end{algorithmic}
\end{algorithm}

Overall, the Order Frontend and Order Mixing modules mainly serve as front-end components for input feature representation and local order enhancement. Their output,
$\mathbf{x}_{\mathrm{order}}^{\mathrm{mix}}$, is used as the unified input to the subsequent shared second-order vibration kernel, MIMO coupling, and adaptive physical correction modules. Since the main focus of this work is on the physics-driven branch and its structured correction mechanism, these two modules are only briefly described here as necessary front-end processing modules.

\subsection{Shared SOS Kernels}

As shown in Part D of Fig.~\ref{fig:model_structure}, PD-SOVNet adopts a dual-branch architecture, where the upper branch is the physics-driven branch and the lower branch is the time-domain branch. The core design of this work lies in the physics-driven branch, which can be further divided into two parts: the base physical branch and the adaptive physical correction branch. The starting point of the base physical branch is the Shared SOS Kernels shown in Parts A and C of Fig.~\ref{fig:model_structure}. The role of this module is not to directly generate the final prediction, but to construct a set of modal response bases with explicit physical meaning in the order domain, so as to explicitly describe the amplification, attenuation, and resonance behaviors that may occur around different orders in the wheel--rail vibration system. These basic modal responses are then combined with MIMO coupling and sample-adaptive correction to form the final output of the physical branch.

\begin{figure*}[t]
    \centering
    \includegraphics[width=1\textwidth]{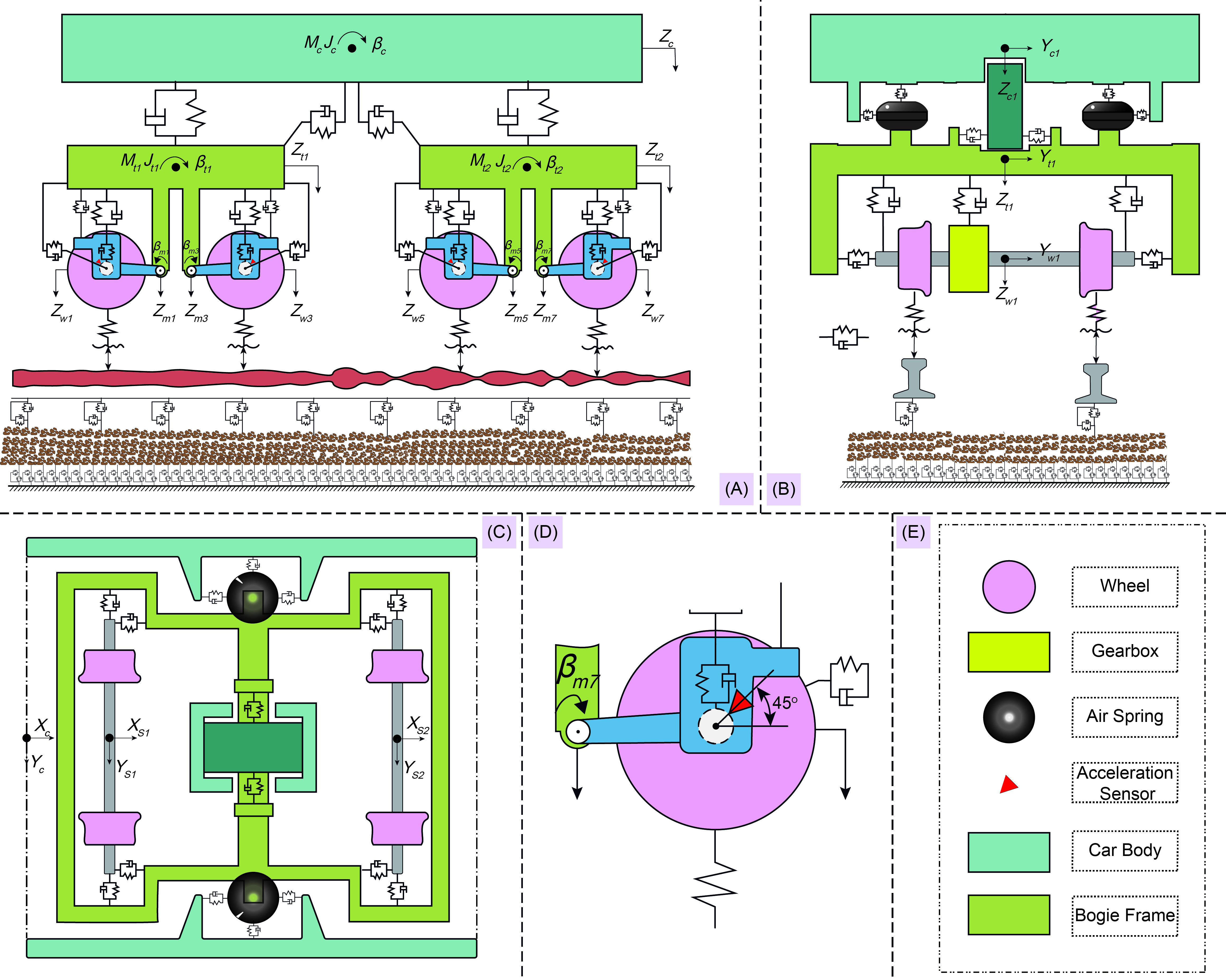}
    \caption{Schematic illustration of the vehicle--bogie--wheelset--rail system. Panels A--C show the multi-level coupled vibration relationships among the carbody, bogie frame, wheelset, and rail. Panel D shows the sensor placement in the axle-box region, indicating that the measured vibration signal is essentially a composite response of wheel--rail excitation after transmission through multiple structural levels.}
    \label{fig:car_structure}
\end{figure*}

This design is not a purely empirical network construction, but is physically motivated by the coupled vibration process of the vehicle--bogie--wheelset--rail system. As shown in Panels A--C of Fig.~\ref{fig:car_structure}, vibration transmission in the vehicle system is jointly determined by the multi-level spring--damper coupling relationships among the carbody, bogie frame, wheelset, and rail, and the local response can, to some extent, be approximately represented by a combination of several second-order vibration units. Meanwhile, the interactions among wheelsets as well as between the wheel and rail further introduce additional coupling and mixing effects. Based on this understanding, the part of the complex wheel--rail system related to modal responses is abstracted in this work as a set of shared second-order vibration kernels, which are used to describe the basic amplification, attenuation, and resonance behaviors around different orders, while the more complex cross-channel transmission and coupling relationships are modeled by the subsequent MIMO coupling module. In other words, Shared SOS Kernels do not attempt to fully reproduce all dynamic details of the vehicle system, especially the wheel--rail coupling part, whose limitation will be discussed later, but instead serve as a physically meaningful low-dimensional modal prior for the structured approximation of the original complex vibration process. Furthermore, the sensor placement shown in Panel D of Fig.~\ref{fig:car_structure} also indicates that the modeled input signal is essentially a composite response of wheel--rail excitation after being transmitted through the wheelset, bogie frame, and suspension system, which further supports the rationality of abstractly characterizing its dominant dynamic features using modal response kernels.

From the modeling perspective, Shared SOS Kernels are not convolution kernels in the conventional sense, but rather a set of \emph{second-order modal response kernels defined on the discrete order domain}. For a single vibration mode in the wheel--rail system, if a standard second-order system is adopted, its continuous-time transfer function can be written as
\[
H_m(s)=\frac{K_m\omega_{n,m}^{2}}{s^{2}+2\zeta_m\omega_{n,m}s+\omega_{n,m}^{2}},
\]
where $K_m$ denotes the gain of the $m$-th mode, $\omega_{n,m}$ denotes the natural frequency, and $\zeta_m$ denotes the damping ratio. Letting $s=j\omega$, the corresponding frequency-response magnitude is
\[
|H_m(j\omega)|
=
\frac{K_m}{
\sqrt{
\left(1-\left(\frac{\omega}{\omega_{n,m}}\right)^2\right)^2
+
\left(2\zeta_m\frac{\omega}{\omega_{n,m}}\right)^2
}
}.
\]

For the wheel polygonization problem, however, the more natural analysis variable is not the absolute frequency, but the order associated with the wheel rotation period. Therefore, this work maps the above second-order-system response from the frequency domain into the order domain and constructs modal response kernels on discrete order sampling points. Let the target order grid be
$\mathcal{O}=\{o_k\}_{k=1}^{40}$,
where $o_k$ denotes the $k$-th order point, and $o$ is the abbreviation of \emph{order}. Furthermore, let the center order, damping, and gain of the $m$-th mode be denoted by $c_m$, $\zeta_m$, and $g_m$, respectively (shown as $Center$, $Damp$, and $Gain$ in the figure for convenience). Then, the discrete response kernel of this mode in the order domain can be written as
\[
\mathbf{k}_m[o_k]
=
\frac{g_m}{
\sqrt{
\left(1-\left(\frac{o_k}{c_m}\right)^2\right)^2
+
\left(2\zeta_m\frac{o_k}{c_m}\right)^2
+\varepsilon
}
},
\qquad k=1,\ldots,40,
\]
where $\varepsilon$ is a small constant introduced to avoid numerical instability. In this way, each mode generates one response curve over the 40 discrete order points, reflecting its selective amplification and attenuation characteristics around different orders.

In this work, Shared SOS Kernels are modeled using a fixed number of modal kernels. Let the number of modes be $M$, where $M=12$ is used in this study. Then each mode is controlled by three learnable parameters, namely
$\boldsymbol{\theta}_m=(c_m,\zeta_m,g_m)=(Center,Damp,Gain),\ m=1,\ldots,M.$
By stacking the response kernels of all modes over the 40 order points, the base kernel bank can be obtained as
\[
\mathbf{K}_{\mathrm{base}}
=
\left[
\mathbf{k}_1,\mathbf{k}_2,\ldots,\mathbf{k}_M
\right]^{\top}
\in\mathbb{R}^{M\times 40}.
\]
Therefore, from the parameter perspective, Shared SOS Kernels essentially implement a mapping from $[M,3]\rightarrow[M,40]$, that is, from the three physical parameters of each mode to its discrete response curve over the 40 target orders. This exactly corresponds to the construction process from modal parameters to order response curves shown in Part A of Fig.~\ref{fig:model_structure}.

The meaning of the term ``shared'' is that the above modal kernel bank is shared across all samples in the base physical branch and is used to describe relatively stable basic vibration priors that are independent of specific samples. In other words, Shared SOS Kernels first provide a set of globally shared modal response templates, while the shifts caused by wheel condition, operating condition, and transmission-path variations across different samples are further modeled by the subsequent adaptive physical correction module. This shared-first, corrected-later structure can, on the one hand, avoid learning the entire modal structure completely from sample-driven signals, and, on the other hand, preserve sufficient flexibility for subsequent context-adaptive correction.

In the subsequent process, given the mixed order representation obtained from the front-end module,
$\mathbf{x}_{\mathrm{order}}^{\mathrm{mix}}\in\mathbb{R}^{40\times4}$,
Shared SOS Kernels do not directly perform channel-wise convolution with the input. Instead, they first generate $M$ groups of shared modal kernels, which are then combined mode by mode with the MIMO coupling results in the subsequent base physical branch. Let the $4\times4$ coupling matrix corresponding to the $m$-th mode be denoted by $\mathbf{W}_m$, labeled as Base W in the figure. Then its modal response can be expressed as
\[
\mathbf{z}_m
=
\mathbf{x}_{\mathrm{order}}^{\mathrm{mix}}\mathbf{W}_m^{\top},
\qquad
\mathbf{z}_m\in\mathbb{R}^{40\times4},
\]
and by weighting it order by order using the $m$-th modal kernel, one obtains
\[
\mathbf{y}_m
=
\mathbf{k}_m\odot \mathbf{z}_m,
\qquad
\mathbf{y}_m\in\mathbb{R}^{40\times4},
\]
where $\odot$ denotes pointwise broadcast multiplication along the order dimension. Finally, summing all modal responses yields the base physical output
\[
\mathbf{y}_{\mathrm{base}}
=
\sum_{m=1}^{M}\mathbf{y}_m.
\]
In this way, Shared SOS Kernels provide the most fundamental ``modal basis functions'' of the physical branch, while MIMO coupling is responsible for describing the transmission and mixing relationships among the four wheel channels. Together, they constitute the base physical branch shown in Part C of Fig.~\ref{fig:model_structure}.

Overall, the key significance of Shared SOS Kernels lies in that they map the learnable modal center--damping--gain parameters into modal response kernels on the discrete order domain, thereby explicitly injecting the structural prior of a second-order vibration system into the network. Compared with directly fitting the input--output mapping using a black-box network, this design enables the model to first learn a set of basic modal templates with explicit physical meaning, and then let the subsequent modules correct the sample-related shifts and coupling variations. Therefore, Shared SOS Kernels are not only the starting point of the base physical branch, but also the core foundation for realizing physics-driven modeling in this work.
\subsection{MIMO Coupling}

The \emph{Base W} introduced in the Shared SOS Kernels section is further elaborated here. As shown in Part D of Fig.~\ref{fig:model_structure}, after the Order Frontend and Order Mixing modules, the model obtains the mixed order representation $\mathbf{x}_{\mathrm{mix}}\in\mathbb{R}^{40\times4}$, which is then fed into the base physical branch. The core objective of this branch is not to directly perform a black-box mapping on the input, but to combine the shared modal response kernels with cross-channel coupling relationships, so as to construct a set of modal responses with explicit structural meaning in the order domain, and then obtain the base physical output through modal summation. Corresponding to Part C of Fig.~\ref{fig:model_structure}, this process mainly consists of four consecutive steps: MIMO coupling, modal mixing, modal weighting, and modal summation.

First, for each mode, a shared MIMO coupling matrix is constructed to describe the basic transmission relationships among the four wheel channels. Unlike traditional FRF-based modeling, in which symmetry or specific physical structural constraints are often explicitly imposed in parameterization, a simpler shared coupling structure is adopted here. For the $m$-th mode, the coupling matrix is defined as
\[
\mathbf{W}_m=\mathrm{Diag}(\mathbf{d}_m)+\mathbf{U}_m\mathbf{V}_m^{\top},
\qquad 
m=1,2,\ldots,M,
\]
where $\mathbf{d}_m\in\mathbb{R}^{4}$ controls the main diagonal response strength of the four channels, and $\mathbf{U}_m,\mathbf{V}_m\in\mathbb{R}^{4\times r}$ are low-rank factors, with $r$ denoting the low-rank dimension. In this way, the $m$-th mode corresponds to a $4\times4$ coupling matrix, and all modes together form the base coupling bank $\mathbf{W}_{\mathrm{base}}\in\mathbb{R}^{M\times4\times4}$. This structure has two characteristics: first, the diagonal terms preserve the independent response capability of each channel; second, the low-rank term provides cross-channel mixing capability while avoiding the parameter redundancy introduced by directly learning a full free-form matrix.

The subsequent part has already been described in the Shared SOS Kernels section. In summary, the MIMO coupling and modal summation process can be summarized in Algorithm~\ref{alg:mimo_modal_sum}.

\begin{algorithm}[t]
\caption{MIMO Coupling and Modal Summation}
\label{alg:mimo_modal_sum}
\begin{algorithmic}[1]
\Require mixed order representation $\mathbf{x}_{\mathrm{mix}}$, shared modal kernels $\{\mathbf{k}_m\}_{m=1}^{M}$, modal coupling matrices $\{\mathbf{W}_m\}_{m=1}^{M}$
\Ensure base physical response $\mathbf{y}_{\mathrm{base}}^{\mathrm{raw}}$
\For{$m=1,2,\ldots,M$}
    \State Construct the coupling matrix $\mathbf{W}_m=\mathrm{Diag}(\mathbf{d}_m)+\mathbf{U}_m\mathbf{V}_m^{\top}$
    \State Compute the modal coupling output $\mathbf{z}_m=\mathbf{x}_{\mathrm{mix}}\mathbf{W}_m^{\top}$
    \State Apply order-wise weighting using the shared SOS kernel to obtain $\mathbf{y}_m=\mathbf{z}_m\odot\mathbf{k}_m$
\EndFor
\State Sum the outputs of all modes to obtain $\mathbf{y}_{\mathrm{base}}^{\mathrm{raw}}$
\end{algorithmic}
\end{algorithm}

Overall, the key idea of this module is to first use the shared MIMO bank to establish modal-level channel transmission relationships for the four-channel input, then apply order-selective weighting using the shared SOS kernels, and finally obtain the base physical response through modal summation. Compared with directly learning a black-box mapping from input to output, this structured process of ``coupling--weighting--summation'' is, in form, closer to the idea of modal superposition, and also provides a clear reference basis for the subsequent adaptive physical correction.

\subsection{Adaptive Physical Correction Branch}

As shown in Part D of Fig.~\ref{fig:model_structure}, the physical branch of PD-SOVNet does not end at the base physical branch. In order to preserve the shared physical priors while adapting to response differences across different wheels, operating conditions, and samples, this work further introduces an Adaptive Physical Correction branch on top of the base physical branch. This branch corresponds to the lower part of Part C in Fig.~\ref{fig:model_structure}. Its core idea is not to relearn a completely free residual mapping, but to perform \emph{structured correction} on the base modal kernels and base coupling relationships based on the contextual information of the input sample, thereby enhancing the adaptability of the model without destroying the physical backbone.

Given the mixed order representation obtained in the previous stage, $\mathbf{x}_{\mathrm{mix}}\in\mathbb{R}^{40\times4}$, the Adaptive Physical Correction branch first uses a context encoder to extract sample-level contextual features. This encoder jointly considers local patterns in the order domain and statistical features, and compresses the overall response pattern of the current sample into a context vector
\[
\mathbf{z}=\mathcal{E}(\mathbf{x}_{\mathrm{mix}}),
\qquad
\mathbf{z}\in\mathbb{R}^{d_c},
\]
where $\mathcal{E}(\cdot)$ denotes the context encoding mapping and $d_c$ is the context feature dimension. This context vector is then fed into two parallel lightweight prediction heads, which are used to predict modal-parameter corrections and coupling-matrix corrections, respectively.

The first branch is the adaptive modal correction branch. Its objective is to flexibly adjust the modal center, damping, and gain according to the contextual features of the current sample, based on the base modal kernels. Let the base modal parameters be denoted by
$\mathbf{c}\in\mathbb{R}^{M}$,
$\boldsymbol{\zeta}\in\mathbb{R}^{M}$,
and
$\mathbf{g}\in\mathbb{R}^{M}$,
respectively. The modal correction head then outputs the corresponding offsets
$\Delta\mathbf{c}$,
$\Delta\log\boldsymbol{\zeta}$,
and
$\Delta\log\mathbf{g}$.
Accordingly, the sample-adaptive modal parameters can be constructed as
\[
\tilde{\mathbf{c}}=\mathbf{c}+\Delta\mathbf{c},
\qquad
\tilde{\boldsymbol{\zeta}}=\boldsymbol{\zeta}\odot \exp(\Delta\log\boldsymbol{\zeta}),
\qquad
\tilde{\mathbf{g}}=\mathbf{g}\odot \exp(\Delta\log\mathbf{g}).
\]
Using these adaptive modal parameters, a sample-dependent adaptive modal kernel bank
$\mathbf{K}_{\mathrm{adp}}\in\mathbb{R}^{M\times40}$
is then regenerated. It should be emphasized that this step does not directly relearn a completely independent new kernel bank. Instead, it \emph{reuses the base modal kernels} and constructs differential corrections on top of them. The design principle is inspired by Differential Transformer \cite{ye2025differential}, which draws on the idea of differential amplifiers to suppress prediction errors caused by noise. Let the base kernel bank be
$\mathbf{K}_{\mathrm{base}}\in\mathbb{R}^{M\times40}$.
Then the kernel correction term is written as
\[
\Delta \mathbf{K}
=
\mathbf{K}_{\mathrm{adp}}-\mathbf{K}_{\mathrm{base}}.
\]
The purpose of this design is to constrain the correction to the \emph{offset relative to the base physical prior}, rather than introducing an unconstrained free residual. In other words, the kernel correction part of the Adaptive Physical Correction branch is essentially a structured incremental correction to the existing physical kernels, rather than replacing the original physical kernels with a data-driven branch.

The second branch is the adaptive coupling correction branch. Its objective is to adjust the base MIMO coupling matrices according to the sample context. Let the base coupling bank be
$\mathbf{W}_{\mathrm{base}}\in\mathbb{R}^{M\times4\times4}$.
Then the coupling correction head outputs
$\Delta\mathbf{W}\in\mathbb{R}^{M\times4\times4}$.
Similar to the modal correction, this branch does not directly learn a completely new coupling structure, but instead constructs sample-dependent incremental corrections around the base coupling matrices, so as to describe the changes in cross-channel transmission relationships under different samples. After obtaining the kernel correction term and the coupling correction term, they are organized into two types of structured physical corrections.

First, the kernel correction is constructed using the base coupling result and the kernel correction term:
\[
\mathbf{y}_{\Delta K}
=
\sum_{m=1}^{M}
\left(
\mathbf{z}_m
\odot
\Delta \mathbf{k}_m
\right),
\]
where $\mathbf{z}_m$ denotes the coupling output of the $m$-th mode in the base physical branch, and $\Delta \mathbf{k}_m$ denotes the correction term of the corresponding modal kernel. This term reflects the output variation caused by correcting only the modal kernel shape, center position, and intensity while keeping the base channel mixing relationship unchanged.

Second, the coupling correction is constructed using the coupling correction term and the adaptive modal kernels:
\[
\mathbf{y}_{\Delta W}
=
\sum_{m=1}^{M}
\left(
\Delta\mathbf{W}_m\mathbf{x}_{\mathrm{mix}}^{\top}
\right)^{\top}
\odot
\tilde{\mathbf{k}}_m,
\]
where $\tilde{\mathbf{k}}_m$ is the adaptive kernel of the $m$-th mode. This term reflects the additional output correction caused by changes in the coupling relationships while keeping the adaptive modal kernels fixed.

By combining the above two structured corrections, the total output of the correction branch can be obtained as
\[
\mathbf{y}_{\Delta}
=
\mathbf{y}_{\Delta K}
+
\mathbf{y}_{\Delta W}.
\]
Finally, this correction term is not directly added to the base physical branch without constraint. Instead, a learnable gating coefficient $\beta$ is introduced to control its contribution:
\[
\mathbf{y}_{\mathrm{phys}}
=
\mathbf{y}_{\mathrm{base}}
+
\beta\,\mathbf{y}_{\Delta}.
\]
Here, $\beta$ is obtained by applying a Sigmoid mapping to a learnable parameter and is constrained to a relatively small range. Its initial value is set to 0, so as to prevent $\mathbf{y}_{\Delta}$ from becoming too strong at the beginning of training before the pure physical branch has learned meaningful content. The function of this gating design is to ensure that the correction branch always serves as a supplement to the base physical branch, rather than evolving into the dominant term during training. In other words, the purpose of Adaptive Physical Correction is not to overpower the base physical modeling, but to perform controlled correction of sample-related local deviations on the basis that the physical backbone has already provided a reasonable response structure, thereby improving representational capacity while maintaining physical consistency.

The above process can be summarized in Algorithm~\ref{alg:adaptive_correction}.

\begin{algorithm}[t]
\caption{Adaptive Physical Correction}
\label{alg:adaptive_correction}
\begin{algorithmic}[1]
\Require mixed order representation $\mathbf{x}_{\mathrm{mix}}$, base modal kernels $\mathbf{K}_{\mathrm{base}}$, base coupling bank $\mathbf{W}_{\mathrm{base}}$
\Ensure corrected physical response $\mathbf{y}_{\mathrm{phys}}$
\State Extract the sample context feature $\mathbf{z}$ using the context encoder
\State Predict $\Delta\mathbf{c}$, $\Delta\log\boldsymbol{\zeta}$, and $\Delta\log\mathbf{g}$ using the modal correction head
\State Generate the adaptive modal kernels $\mathbf{K}_{\mathrm{adp}}$ based on the corrected modal parameters
\State Compute the kernel correction term $\Delta\mathbf{K}=\mathbf{K}_{\mathrm{adp}}-\mathbf{K}_{\mathrm{base}}$
\State Predict the coupling increment $\Delta\mathbf{W}$ using the coupling correction head
\State Construct the kernel correction based on $\Delta\mathbf{K}$, and construct the coupling correction based on $\Delta\mathbf{W}$
\State Sum the two correction terms to obtain $\mathbf{y}_{\Delta}$
\State Constrain the correction strength by the gating coefficient $\beta$, and add it to the base physical output to obtain $\mathbf{y}_{\mathrm{phys}}$
\end{algorithmic}
\end{algorithm}

Overall, the key of the Adaptive Physical Correction branch does not lie in introducing a stronger free residual, but in constructing a constrained structured correction mechanism around the base physical prior. First, the kernel correction explicitly reuses the base modal kernels, so that the correction is always centered around the existing physical response. Second, the coupling correction only performs sample-related adjustment on the base transmission relationships, rather than redefining the entire coupling structure. Third, the gating term $\beta$ further ensures that the correction term always remains auxiliary. Through this design, the model can more flexibly adapt to modal shifts and coupling variations under different samples and operating conditions, without weakening the interpretability of the base physical modeling.

\subsection{Time-Domain Branch and Branch Fusion}

\begin{figure}
    \centering
    \includegraphics[width=1\linewidth]{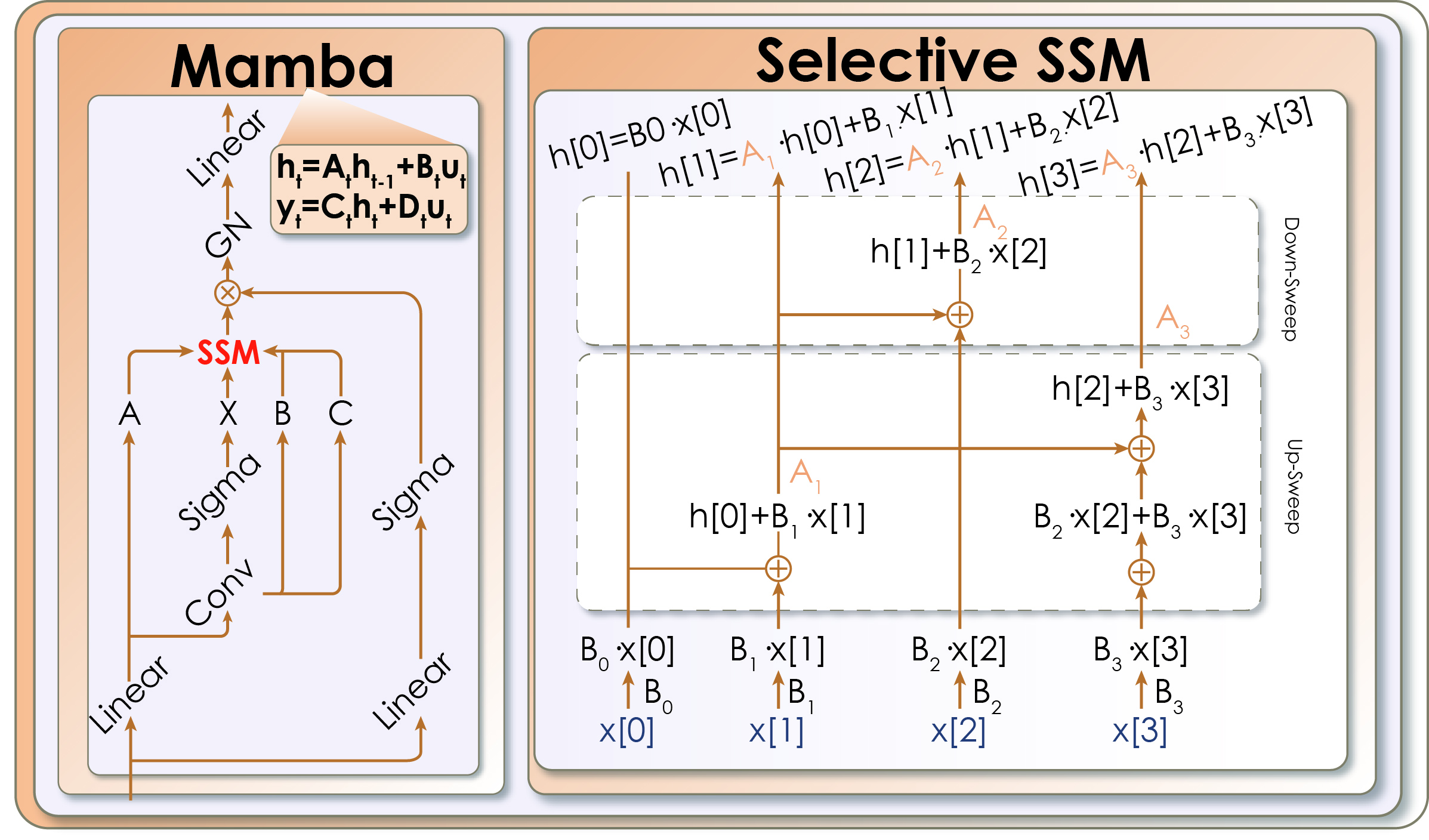}
    \caption{Mamba structure and selective SSM.}
    \label{fig:mambar}
\end{figure}

In addition to the physics-driven branch, this work also introduces a parallel time-domain branch, which is used to supplement the extraction of dynamic information from the raw angle-domain vibration sequence that is difficult to be fully described by explicit physical priors. Given the input
$\mathbf{x}\in\mathbb{R}^{400\times4}$ and
$\mathbf{v}\in\mathbb{R}^{4}$,
the time-domain branch first linearly embeds the vibration input and the speed information separately, then injects the speed features into the temporal representation, and further models long-range dependencies through a two-layer Mamba structure (see Fig.~\ref{fig:mambar}). Finally, the time-domain features are projected from the angular-point dimension into the target order space, yielding the output of the time-domain branch,
$\mathbf{y}_{\mathrm{time}}\in\mathbb{R}^{40\times4}$.
The role of this branch is mainly to complement the residual temporal dynamics that cannot be fully covered by the physical branch. Therefore, it is not discussed as a major innovation module in this work.

After obtaining the physical-branch output
$\mathbf{y}_{\mathrm{phys}}\in\mathbb{R}^{40\times4}$
and the time-domain-branch output
$\mathbf{y}_{\mathrm{time}}\in\mathbb{R}^{40\times4}$,
the model performs branch fusion using a learnable gating coefficient. Let the fusion coefficient be
$\alpha\in(0,1)$.
Then the linear fusion result can be written as
\[
\mathbf{y}_{\mathrm{lin}}
=
(1-\alpha)\mathbf{y}_{\mathrm{phys}}
+
\alpha \mathbf{y}_{\mathrm{time}},
\qquad
\alpha=\sigma(a),
\]
where $\sigma(\cdot)$ denotes the Sigmoid function and $a$ is a learnable scalar parameter. This design enables the model to adaptively balance the contributions of the physical branch and the time-domain branch during training.

Since the supervision labels in this work are roughness spectra represented in dB form after frequency-domain transformation during preprocessing, the final output is also constructed in the dB domain, so as to maintain consistency with the label space and reduce the influence of amplitude dynamic range on regression stability. Specifically, the model first maps the fused output to the positive domain through Softplus, and then performs a logarithmic transformation to obtain the final prediction
\[
\hat{\mathbf{y}}
=
20\log_{10}
\left(
\frac{\mathrm{softplus}(\mathbf{y}_{\mathrm{lin}})+\varepsilon}{y_{\mathrm{ref}}}
\right),
\]
where $\varepsilon$ is a numerical stabilization term and $y_{\mathrm{ref}}$ is the reference quantity. Through the $20\log_{10}$ mapping, the model output is represented in the same dB form as the roughness labels, which facilitates the subsequent supervised regression training.

\section{Experiments}

\subsection{Experimental setup}

This section introduces the datasets, data splitting strategy, comparison methods, implementation details, computational characteristics, and evaluation metrics used in the experiments. Since the main focus of this work is to evaluate the roughness-regression capability of the model under unseen-wheel conditions, the experiments do not adopt a simple random split. Instead, the training and validation sets are divided based on wheel groups, so as to better reflect the cross-wheel generalization scenario encountered in practical applications.

It should be noted that this work does not consider similar validation results at an \emph{insufficiently optimized} training stage to be sufficient for fully reflecting the practical value of a model. For a quantitative prediction task such as wheel polygon roughness regression, the more critical issue is whether the model can still maintain strong generalization performance on unseen wheel groups after it has been further optimized to an engineeringly usable error range.

\subsubsection{Datasets and data splitting}

\begin{figure}
    \centering
    \includegraphics[width=0.5\linewidth]{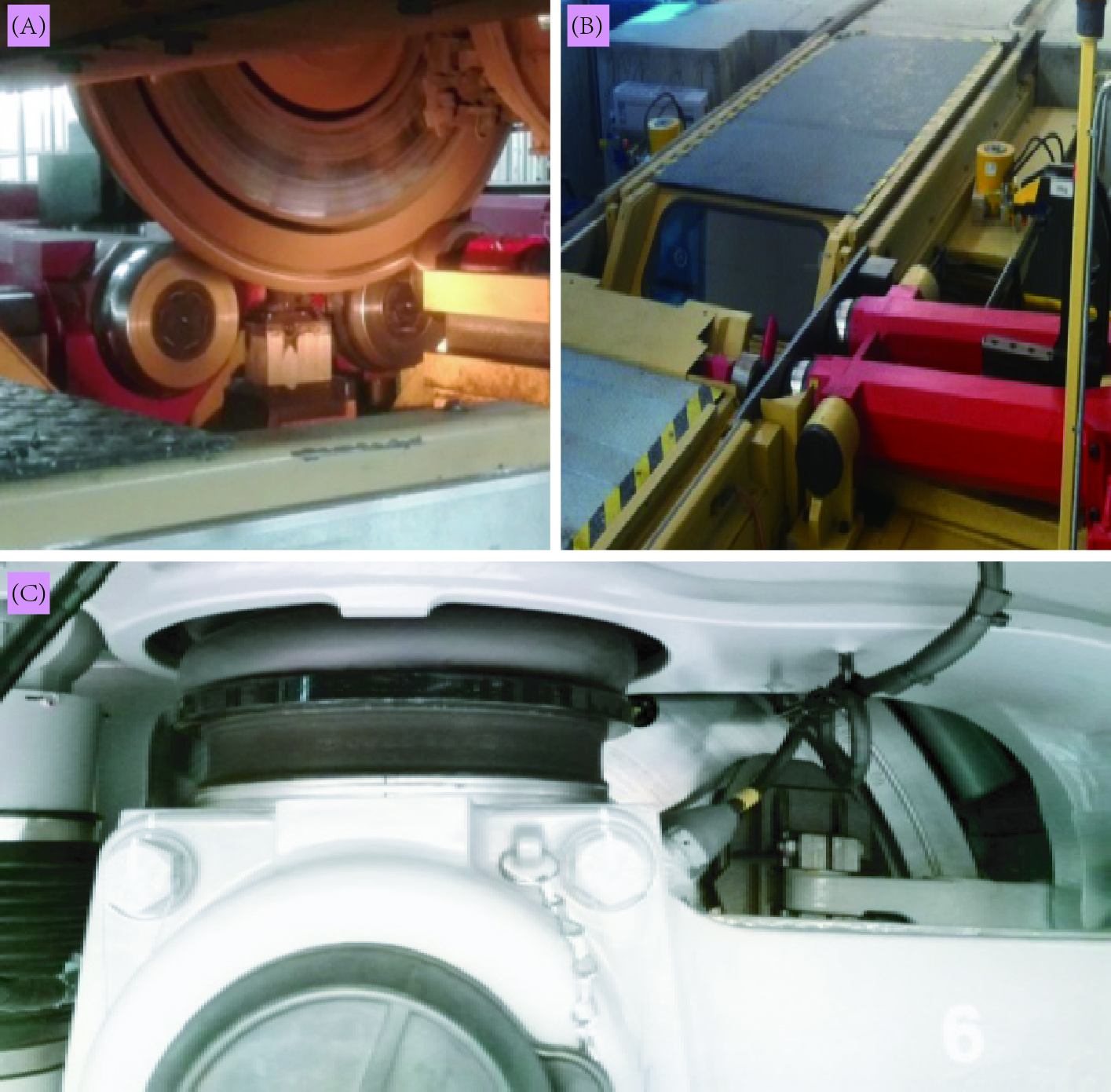}
    \caption{Field collection of wheel roughness and axle-box vibration data. (A) Wheel roughness data acquisition process. (B) Data acquisition equipment. (C) Measurement location of axle-box vibration and on-site sensor installation.}
    \label{fig:threepictureofdatacolletion}
\end{figure}

The effectiveness of the proposed method is validated on three datasets, denoted as Dataset I, Dataset II, and Dataset III, respectively. The data sources can be found in the provided data-collection video, in Step 1 of Fig.~\ref{fig:data_pipeline}, in Fig.~\ref{fig:threepictureofdatacolletion}, or in Section~\ref{数据预处理}. In total, the datasets contain 448 groups of data collected from high-speed railway lines including Beijing--Guangzhou, Beijing--Shanghai, and Jinan--Qingdao, and all trains share a unified vehicle architecture. Among them, the validation-set data of the three datasets are all provided by the data supplier. Dataset I and Dataset II use validation sets formed by two bogies with relatively large differences, corresponding to a total of 8 wheel groups, including two relatively normal data groups. Dataset III uses fault data from the available dataset as the validation set, which are extremely valuable real fault data. More details can be found at the top of Part A in Fig.~\ref{fig:resultof1}. All three datasets adopt the same supervised regression setting, namely inferring wheel polygon roughness values from axle-box vibration responses. For each sample, the input consists of single-revolution angle-domain vibration sequences of four wheels together with the corresponding speed information, and the label is the roughness value of the four wheels at 40 orders.

Specifically, the cached input tensor is represented as
$\mathbf{X}_{4w}\in\mathbb{R}^{N\times 400\times 4}$,
where \(N\) denotes the number of samples after resampling, 400 denotes the number of angular points after single-revolution resampling, and 4 denotes the four wheel channels within the same wheel group. The corresponding labels are denoted by
$\mathbf{Y}_{4w}\in\mathbb{R}^{N\times 40\times 4}$,
and the speed information is denoted by
$\mathbf{V}_{4w}\in\mathbb{R}^{N\times 4}$.

To avoid information leakage caused by having samples from the same wheel group appear simultaneously in both the training set and the validation set, this work adopts a wheel-group-based split rather than a random split. Specifically, the training set and validation set are divided according to wheel groups, that is, some wheel groups are entirely assigned to the training set, while the remaining unseen wheel groups are assigned to the validation set. This splitting strategy can more realistically reflect the generalization capability of the model under unseen-wheel conditions. All experiments in this work follow the same data protocol.

\subsubsection{Comparison methods}

To comprehensively evaluate the effectiveness of the proposed PD-SOVNet, it is compared with five methods, including four purely data-driven baseline models and one physics-based baseline model. The four purely data-driven baselines are CNN, BiLSTM, Mamba, and Transformer, while the physics-based baseline is the FRF model \cite{frf}. The proposed method is denoted as Ours.

To evaluate the proposed method from an application-oriented perspective, this work adopts two complementary analyses instead of relying on a single all-purpose comparison protocol. The second analysis is introduced only as a supplementary experiment to illustrate the difference between the deployment-oriented performance of the proposed method and the FRF-based model, and the regularized performance of the other four data-driven models. To address the potential concern that this setting may not constitute a strictly matched comparison, we have further supplemented the study in Sec. \ref{sec:FRFouradd}, where the complete comparison results are fully presented.s

The first analysis is a \textbf{primary low-training-error comparison} designed to reflect the current deployment objective. Under this analysis, all methods follow the same data split, training budget, and weight-selection rule, and the checkpoint is selected according to a unified low-training-error criterion. Therefore, this analysis should be interpreted as the main comparison under the current protocol rather than merely as an observational stability analysis.

The second analysis is a \textbf{supplementary training-state sensitivity analysis}. By comparing a regularized training state with a further-optimized training state, this analysis is used to examine how different methods respond to training-state changes under the unseen-wheel setting. Its role is to provide additional insight into stability trends, not to serve as an independent basis for absolute ranking claims. For readers to see the complete comparison information, the supplementary experiment in Section~\ref{sec:FRFouradd} further provides the fair comparison between the two training schemes together with the remaining related content.

Accordingly, the results in this section are interpreted mainly in terms of the main comparison under the current protocol, supplementary sensitivity trends across training states, and engineering relevance under the current data protocol.

\subsubsection{Implementation details}

All models are trained and evaluated using the same input format, where the input consists of four-channel single-revolution angle-domain vibration sequences and their corresponding speed information, and the output is the predicted roughness values of the four wheels at 40 orders. All models are uniformly trained for 30 epochs (the reported results are the averages over 5 runs), and early stopping is not adopted, including in the regularized training state, so as to ensure consistency in the number of training epochs across different methods. AdamW is used as the optimizer for all models. For the four purely data-driven baselines under the regular training state, dropout and regularization strategies are introduced to reduce excessive fitting during training.

The learning rate and batch size are kept at the same order of magnitude across different models, so as to minimize the influence of training-hyperparameter differences on the results as much as possible. In addition, one set of model weights is specifically retained when the training MAE loss first drops to the 0.8 level, which is later used for the simulated deployment noise-interference experiments.

\subsubsection{Computational characteristics and deployment implications}

To complement the accuracy-oriented evaluation, we further report a supplementary deployment-oriented profile of the compared models under the same measurement pipeline. In practical industrial applications, model adoption depends not only on regression accuracy but also on model compactness, inference latency, throughput, and memory usage. Therefore, these metrics are presented as a supplementary deployment-oriented profile rather than as the primary basis for ranking model quality.

All compared models were evaluated under the same inference pipeline with a unified batch size of 64. Table~\ref{tab:deployment_profile} shows that PD-SOVNet has the smallest parameter count among the compared methods, indicating that the proposed architecture is compact in parameter count. At the same time, under the current implementation it is not the fastest or most memory-efficient model. Accordingly, the computational profile of PD-SOVNet is better described as compact in parameter count but not yet runtime-optimized.

This observation should be interpreted together with the application setting of this study. In the present maintenance workflow, wheel roughness assessment is performed once per day rather than as a continuously running high-frequency online task. Therefore, the current inference speed is sufficient for the present once-per-day maintenance workflow. Under this setting, the main value lies in relatively stable prediction behavior rather than computational superiority under the unseen-wheel protocol, while future work should further optimize runtime efficiency and memory usage for larger-scale deployment.

\begin{table}[t]
    \centering
    \caption{Computational characteristics of the compared models under the same measurement pipeline with a unified batch size of 64. The table reports parameter count, average per-sample inference latency, throughput, and peak memory allocation. These metrics are provided to complement the accuracy-oriented evaluation and to illustrate deployment-related trade-offs, rather than to serve as the primary basis for model ranking.GPU(3080ti)}
    \label{tab:deployment_profile}
    \setlength{\tabcolsep}{6pt}
    \renewcommand{\arraystretch}{1.1}
    \resizebox{\linewidth}{!}{%
    \begin{tabular}{lcccc}
        \toprule
        Model & Params (M) & Avg. latency (ms/sample) & Throughput (samples/s) & Peak memory (MiB) \\
        \midrule
        BiLSTM      & 1.71 & 0.027 & 36758 & 100.5 \\
        CNN         & 1.44 & 0.033 & 46734 & 94.0 \\
        FRF         & 2.32 & 0.030 & 33630 & 40.5 \\
        Transformer & 8.48 & 0.074 & 13655 & 121.4 \\
        Mamba       & 3.07 & 0.092 & 10925 & 128.8 \\
        Ours        & \textbf{0.55} & 0.140 & 7184 & 203.0 \\
        \bottomrule
    \end{tabular}%
    }
\end{table}

\subsubsection{Evaluation metrics and comparison protocol}

This work adopts mean absolute error (MAE) and coefficient of determination \(R^2\) as the main evaluation metrics. MAE is used to measure the average absolute deviation between the model prediction and the ground-truth roughness value, and can directly reflect the magnitude of the regression error. \(R^2\) is used to measure the ability of the model to explain the variation trend of the target, and can evaluate the model performance from the perspective of overall fitting quality. All results are reported separately on Dataset I, Dataset II, and Dataset III, and are presented in the subsequent sections through tables and bar charts.

In addition to the main comparison experiments, this work further designs noise-interference experiments to examine model behavior under perturbed inputs near a low-training-error reference state. Specifically, based on a set of model weights trained to a relatively low error level, noise perturbations of different intensities are added to the input samples, and the degradation trends of different models under increasing noise are compared. This experiment is mainly used to evaluate how the improved physics-guided model behaves under simulated noise. The corresponding ablation experiments provide supplementary evidence that physically constrained models can experience performance degradation under noise interference \cite{noise_1,noise_2}, and that the temporal branch helps alleviate part of this problem.

\subsection{Low-training-error stability analysis}

\begin{figure*}[!htbp]
    \vspace{-80pt}
    \centering
    \includegraphics[width=1\linewidth]{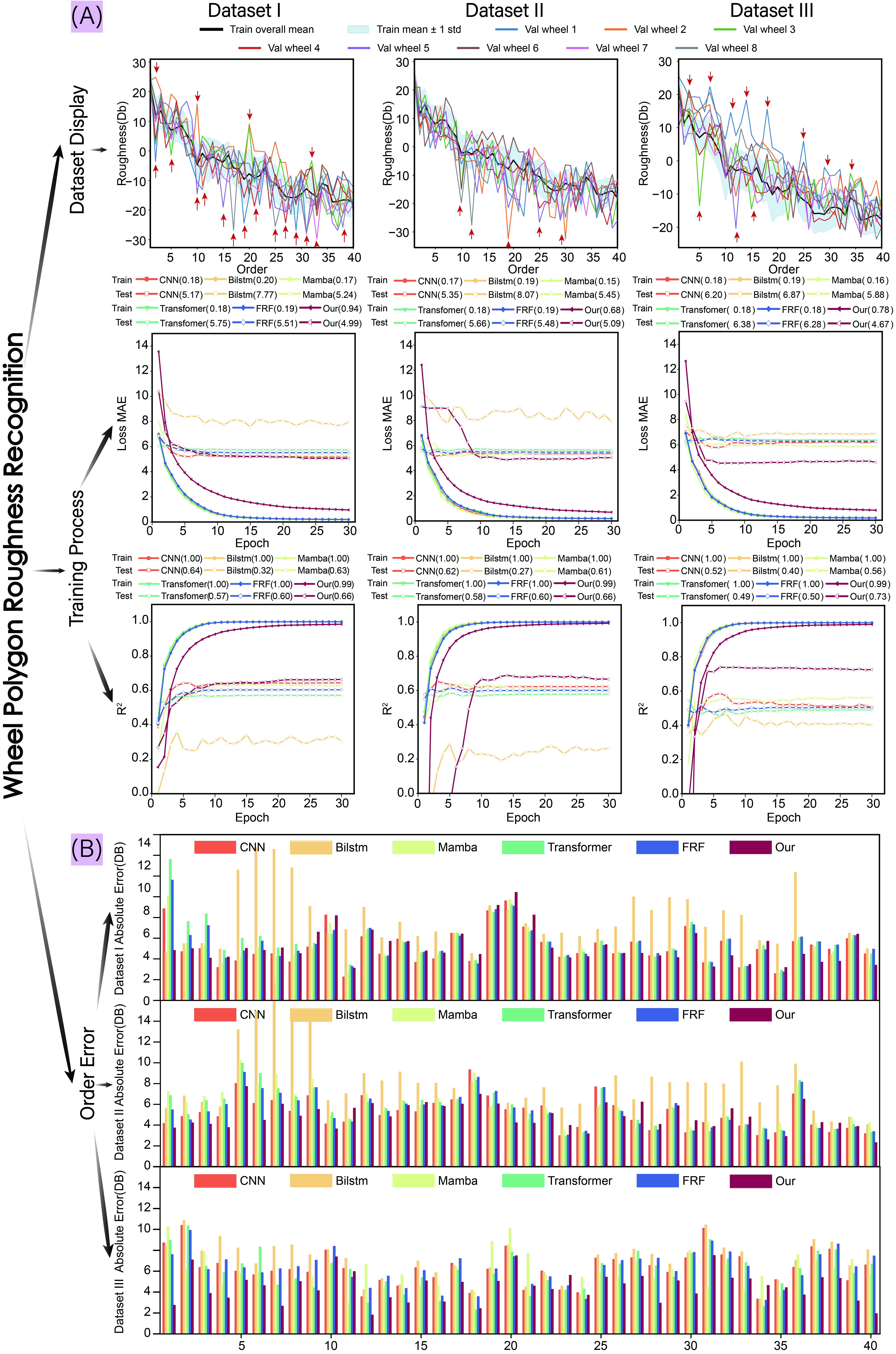}
    \caption{Comparison results of six methods on the three datasets under the low-training-error stability analysis. (A) shows examples of the validation roughness spectra on Dataset I, Dataset II, and Dataset III, together with the curves of MAE (lower is better) and $R^2$ (higher is better) during training for each method. The validation roughness spectra are used to show how different methods fit the target roughness distribution over orders, while the MAE and $R^2$ curves reflect the error convergence and fitting trends during training. (B) shows the absolute error distributions over 40 orders on the three datasets, which are used to compare the error level and stability of different methods in order-wise prediction. This analysis is used mainly to examine performance stability when the models are trained to a lower training-error range under the unseen-wheel split.}
    \label{fig:resultof2}
\end{figure*}

This subsection reports the \textbf{primary low-training-error comparison}. Under this analysis, all methods follow the same data split, training budget, and weight-selection rule, and the checkpoint is selected according to a unified low-training-error criterion. Therefore, this subsection should be interpreted as the main comparison under the current protocol. The emphasis is on validation performance after the models have reached the shared low-training-error target under the unseen-wheel split.

Fig.~\ref{fig:resultof2}(A) shows examples of the validation roughness spectra on Dataset I, Dataset II, and Dataset III, together with the MAE and $R^2$ curves during training for each method. Fig.~\ref{fig:resultof2}(B) further presents the order-wise error distributions over the 40 orders on the three datasets. From the overall trend, PD-SOVNet shows competitive validation performance on all three datasets, with its more noticeable stability advantage appearing on Dataset III. The training curves indicate that several purely data-driven baselines can achieve very low training errors, but their validation behavior is generally more sensitive to the gap between the training and validation distributions. In contrast, although PD-SOVNet does not always achieve the lowest training error, its validation MAE and $R^2$ remain comparatively stable under this analysis. These observations suggest that the proposed gray-box structure may help reduce sensitivity to distribution shift under the current data protocol.

On Datasets I and II, the margins between PD-SOVNet and the stronger baselines are generally moderate rather than large, so the main message here is better understood as relative stability and result consistency rather than absolute dominance. On Dataset III, where the validation distribution appears more challenging, the advantage of PD-SOVNet becomes more visible, which is consistent with the interpretation that structured physical priors may be more helpful when the distribution shift is stronger.

The order-wise error distributions in Fig.~\ref{fig:resultof2}(B) provide a similar picture. Compared with several purely data-driven baselines, the proposed method tends to show more balanced error behavior across orders, while FRF remains a relevant physics-based reference but is generally less stable under this analysis. At the same time, some baselines remain competitive on individual datasets or metrics, so this subsection should be read as the main comparison under the current unseen-wheel protocol, with the focus placed on the full set of results rather than on isolated metric advantages.

\subsection{Training-state sensitivity analysis}

\begin{figure*}[!htbp]
    \vspace{-80pt}
    \centering
    \includegraphics[width=1\linewidth]{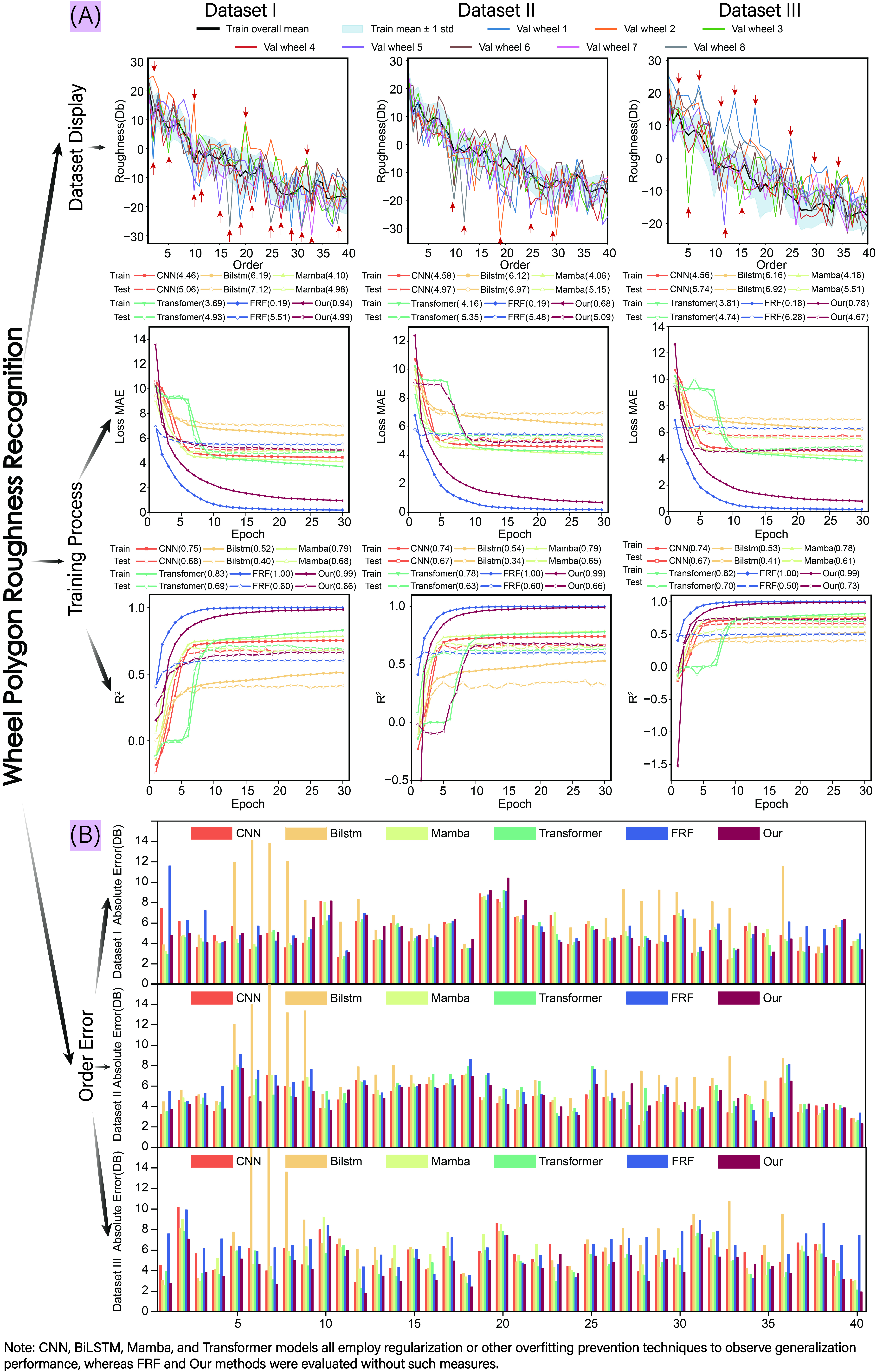}
    \caption{Comparison results of six methods on the three datasets under the training-state sensitivity analysis. (A) shows examples of the validation roughness spectra on Dataset I, Dataset II, and Dataset III, together with the curves of MAE (lower is better) and $R^2$ (higher is better) during training for each method. The validation roughness spectra are used to show how different methods fit the target roughness distribution over orders, while the MAE and $R^2$ curves reflect the error convergence and fitting trends during training. (B) shows the absolute error distributions over 40 orders on the three datasets, which are used to compare the error level and stability of different methods in order-wise prediction. In this analysis, the purely data-driven baselines are regularized, while FRF and Ours are additionally shown in a further-optimized training state.}
    \label{fig:resultof1}
\end{figure*}

This subsection reports the \textbf{supplementary training-state sensitivity analysis}. Under this analysis, the four purely data-driven baselines are regularized to examine their generalization behavior, while FRF and PD-SOVNet are additionally shown in a more fully optimized state. The purpose here is to understand how model performance changes across training states, rather than to use this subsection alone as a definitive basis for absolute ranking claims.

Fig.~\ref{fig:resultof1}(A) shows the validation roughness spectra on the three datasets together with the MAE and $R^2$ curves during training, and Fig.~\ref{fig:resultof1}(B) further presents the order-wise error distributions over 40 orders on the three datasets. Under this analysis, PD-SOVNet behaves as a stable and physically interpretable competitive solution. On Datasets I and II, its performance is generally close to that of the stronger baselines, while on Dataset III the advantage becomes more visible. Accordingly, the main message of this subsection is not absolute numerical superiority on every dataset, but comparatively stable behavior across datasets and training states.

The results on Datasets I and II suggest that several regularized data-driven baselines remain competitive, and in some cases the margins relative to PD-SOVNet are small. On Dataset III, however, the benefit of the proposed gray-box structure becomes easier to observe, which is consistent with the interpretation that physically guided priors may be more helpful when the validation distribution is more challenging under the current unseen-wheel split.

The order-wise error distributions in Fig.~\ref{fig:resultof1}(B) support the same reading. Compared with some purely data-driven baselines, the proposed method generally exhibits more balanced cross-order behavior, while FRF remains informative as a physics-based reference but is more sensitive on the more difficult dataset. Overall, this subsection is best interpreted as a supplementary analysis of training-state sensitivity rather than a stand-alone architectural ranking.

\subsection{Integrated comparison across the two evaluation analyses}

\begin{figure}
    \centering
    \includegraphics[width=1\linewidth]{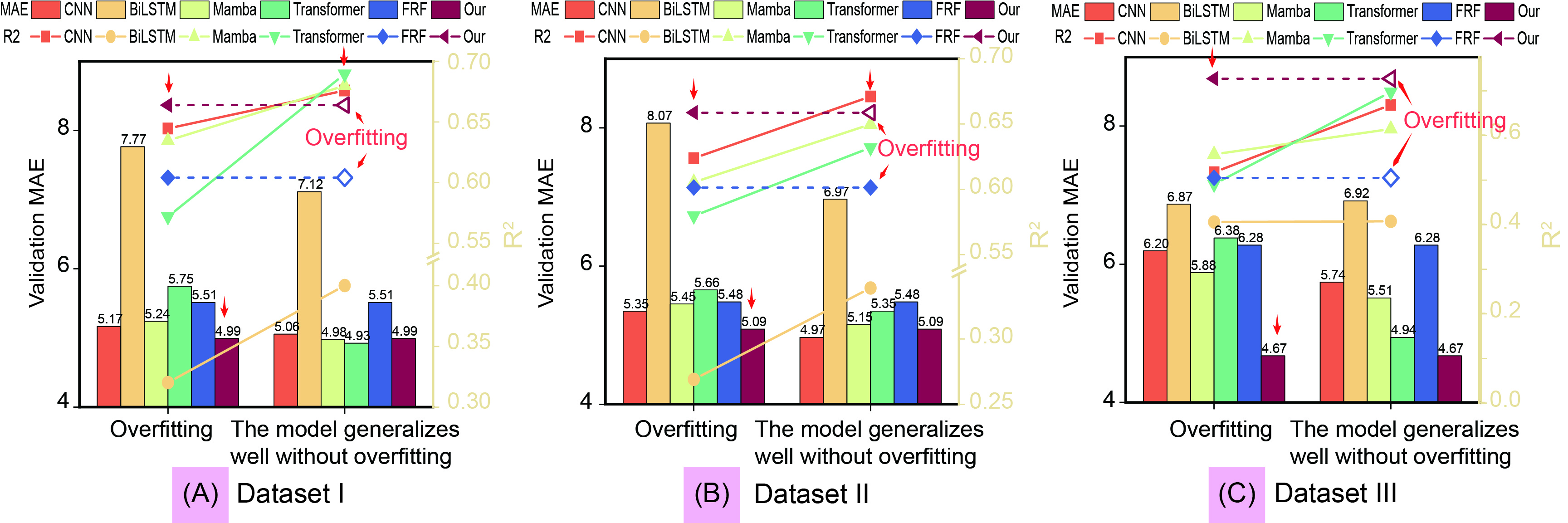}
   \caption{Summary comparison of the final results of different methods on the three datasets under the two evaluation analyses. (A)--(C) correspond to Dataset I, Dataset II, and Dataset III, respectively. In each subfigure, the bar chart represents the validation MAE, and the line plot represents the validation $R^2$; the horizontal axis corresponds to the further-optimized training state and the regularized training state. Here, the term ``overfitting'' in the figure is used only to denote the lower-training-error state after further optimization, rather than as a formal diagnosis of statistical overfitting for all methods.}
    \label{fig:dataanaliese}
\end{figure}

Taken together, the two analyses suggest that the main potential value of PD-SOVNet lies in performance stability under the current unseen-wheel data protocol, rather than in uniformly outperforming every baseline on every dataset and every metric. The benefit becomes more visible when the distribution shift is stronger, particularly on Dataset III, whereas on Datasets I and II the margins against the strongest baselines are often modest.

To facilitate a unified comparison of the results under the two analyses, Fig.~\ref{fig:dataanaliese} further summarizes the final MAE and $R^2$ results of the six methods on the three datasets. From the overall trend, the purely data-driven baselines are more sensitive to the training strategy, and their final results fluctuate more significantly across different settings. In contrast, FRF and PD-SOVNet show relatively more stable behavior across the two analyses, although some baselines remain slightly better on individual metrics and should be interpreted accordingly.

From the summarized results on Dataset I and Dataset II, Ours maintains relatively stable performance under both settings. Specifically, on Dataset I, the MAE of Ours is 4.99\,dB under both settings, which is better than the 5.51\,dB of FRF and also better than BiLSTM; compared with CNN, Mamba, and Transformer, Ours is at a similar or slightly better level. On Dataset II, the MAE of Ours likewise remains at 5.09\,dB, which is slightly better than the 5.48\,dB of FRF and also clearly better than BiLSTM. It should be noted that, under the regularized training setting, CNN achieves an MAE of 4.97\,dB on Dataset II, which is slightly lower than that of Ours; on Dataset I, Transformer achieves an MAE of 4.93\,dB, also slightly better than Ours. This indicates that, Although some regularized data-driven models may obtain results close to, or even slightly better than, the proposed method on certain individual metrics, their training errors fail to meet the required deployment standards, making them unsuitable for practical deployment.

In contrast, the differences are more obvious on Dataset III. Under both the setting where the training error is further reduced and the regularized training setting, the MAE of Ours is 4.67\,dB, which is the lowest among the six methods on this dataset. Correspondingly, the MAE of FRF is 6.28\,dB, that of CNN is 6.20/5.74\,dB, that of Mamba is 5.88/5.51\,dB, that of Transformer is 6.38/4.94\,dB, and that of BiLSTM remains at the relatively high level of 6.87/6.92\,dB. Combined with the variation trend of $R^2$ in the figure, it can be seen that when the distribution gap between the training and validation sets of Dataset III is larger, the purely data-driven models exhibit more obvious fluctuations across different training settings, whereas Ours still maintains a high and stable fitting level. This phenomenon indicates that the proposed method is more likely to show its advantage under more complex data conditions or stronger distribution shifts.

A further comparison among Ours, FRF, and the four purely data-driven baselines shows that the variations of MAE and $R^2$ for FRF and Ours across the three datasets are relatively stable, indicating that physics-related structures can suppress the performance fluctuations caused by changes in training strategy to some extent. However, compared with FRF, Ours achieves lower MAE on all three datasets, especially on Dataset III, which suggests that relying only on fixed physical modeling is still insufficient, and that introducing adaptive physical correction can further improve the expressiveness of the model. Among the four purely data-driven baselines, BiLSTM performs the worst overall; CNN and Mamba perform reasonably well on Dataset I and Dataset II, but their advantages are unstable under more complex data conditions; Transformer is relatively sensitive to the training strategy, and although it can achieve good results on some datasets after regularization, its overall stability is still inferior to that of Ours.

From the perspective of method design, the results reflected in Fig.~\ref{fig:dataanaliese} are consistent with the structural characteristics of the proposed model. PD-SOVNet provides basic modal-response priors in the order domain through shared SOS kernels, characterizes the transmission relationships among the four channels through MIMO coupling, and performs controlled compensation for modal shifts and coupling changes under different samples through adaptive physical correction. Therefore, when the training error is reduced or the training strategy changes, the model does not rely entirely on data fitting to maintain performance, but can instead preserve relatively stable generalization through the physical backbone. In particular, on scenarios such as Dataset III with more obvious distribution differences, the advantage of this gray-box modeling strategy becomes easier to observe.

\subsubsection{Supplementary experiment: completion and integrated presentation of FRF and Ours results}\label{sec:FRFouradd}

Since the main comparison experiments under the training-state sensitivity analysis were primarily intended to show how the methods behave across training states, the complete training-process plots of FRF and the proposed method under this analysis were not separately presented in the previous section. To reduce possible ambiguity in interpretation, we supplement here the further-optimized training processes of FRF and Ours on the three datasets, and further provide a unified summary of the final results of the six methods under two training states.

\begin{figure}[t]
    \vspace{-80pt}
    \centering
    \includegraphics[width=0.3\linewidth]{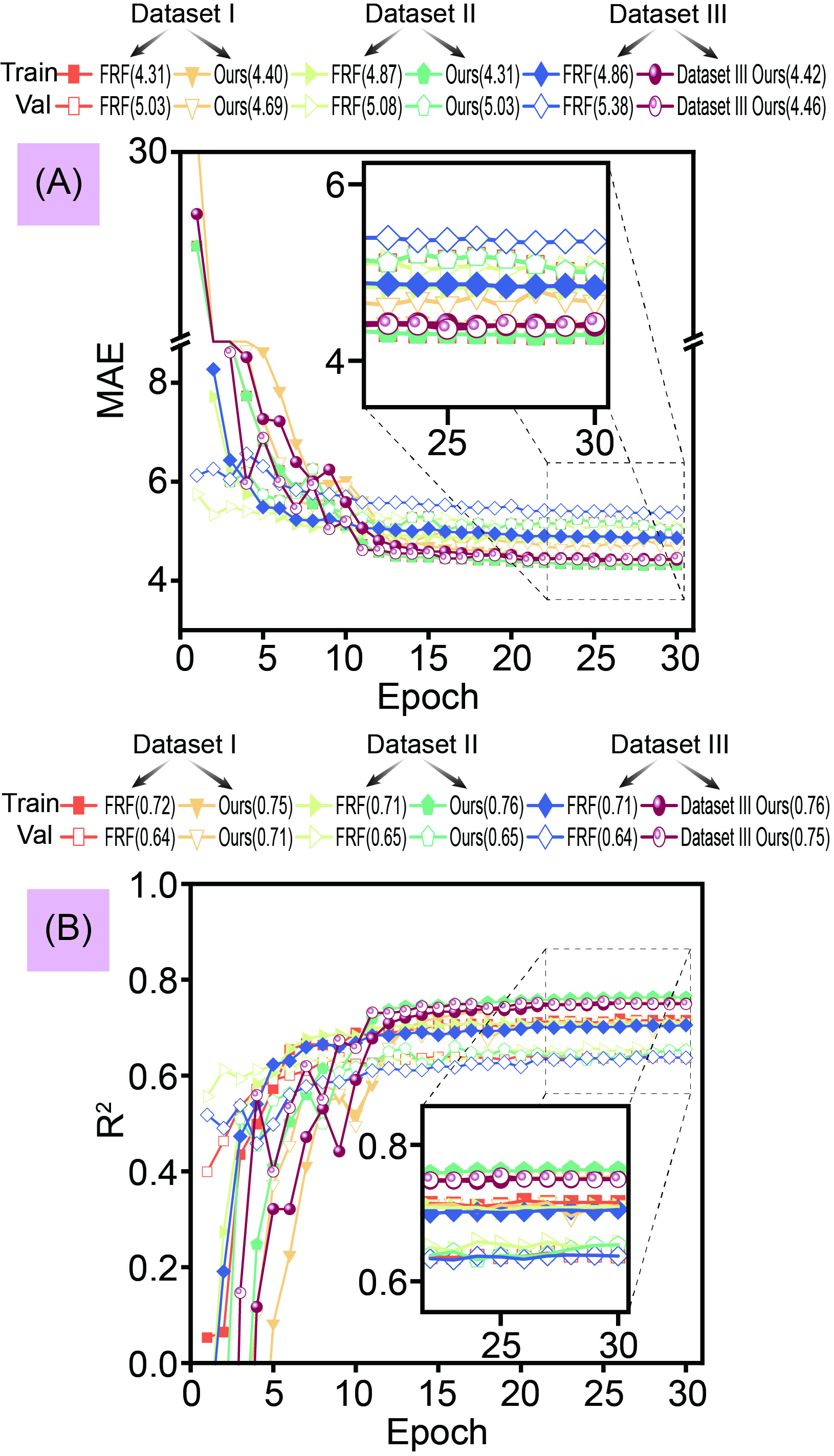}
    \caption{Supplementary training-process results of FRF and Ours on the three datasets under the training-state sensitivity analysis. (A) shows the MAE curves over training epochs; (B) shows the $R^2$ curves over training epochs. This figure is mainly used to supplement the training and validation results of FRF and Ours on Dataset I, Dataset II, and Dataset III in the previous main comparison experiment.}
    \label{fig:datatrainadd}
\end{figure}

\begin{figure}[H]
    \centering
    \includegraphics[width=1\textwidth]{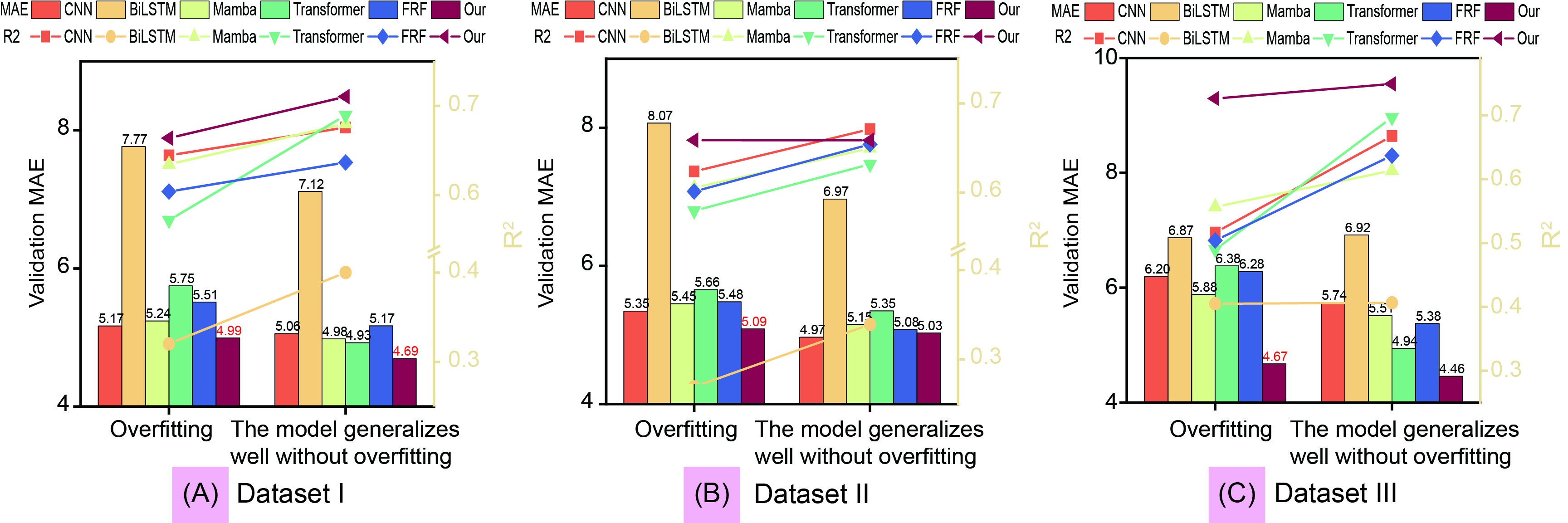}
    \caption{Integrated comparison of the final results of six methods on the three datasets under two training states. (A)--(C) correspond to Dataset I, Dataset II, and Dataset III, respectively. The bar chart represents the validation MAE, and the line plot represents the validation $R^2$; the horizontal axis corresponds to the further-optimized training state and the regular training state. Here, the term ``overfitting'' in the figure is used only to denote the lower-training-error state after further optimization, rather than as a formal diagnosis of statistical overfitting for all methods.}
    \label{fig:contrastnew}
\end{figure}

Figs.~\ref{fig:datatrainadd} and \ref{fig:contrastnew} together provide the supplementary results under the training-state sensitivity analysis. Overall, both FRF and Ours converge relatively quickly on all three datasets, and their curves become stable in the later training stages. Compared with FRF, Ours achieves lower validation MAE on all three datasets, and higher, or at least not lower, validation $R^2$ than the physics-based baseline. Meanwhile, in the integrated results shown in Fig.~\ref{fig:contrastnew}, when switching from the further-optimized training state to the regular training state, the performance changes of Ours are relatively milder, indicating that the proposed method is relatively less sensitive to changes in training state.

For Dataset I, Fig.~\ref{fig:datatrainadd} shows that at the end of training, the validation MAE of Ours is $4.69$, lower than the $5.03$ of FRF; the corresponding validation $R^2$ values are $0.71$ and $0.64$, respectively. The integrated results in Fig.~\ref{fig:contrastnew}(A) further show that under the further-optimized training state, the validation MAE values of CNN, BiLSTM, Mamba, Transformer, FRF, and Ours are $5.17$, $7.77$, $5.24$, $5.75$, $5.51$, and $4.99$, respectively; under the regular training state, they are $5.06$, $7.12$, $4.98$, $4.93$, $5.17$, and $4.69$, respectively. These results suggest that Ours remains competitive on this dataset under both training states, while also showing relatively stable cross-wheel prediction behavior compared with the physics-based baseline and several purely data-driven models.
For Dataset II, Fig.~\ref{fig:datatrainadd} shows that the training processes of Ours and FRF are relatively close, but the final validation MAE of Ours is still further reduced from $5.08$ to $5.03$; both methods achieve a validation $R^2$ of $0.65$. From Fig.~\ref{fig:contrastnew}(B), under the further-optimized training state, the validation MAE values of the six methods are $5.35$, $8.07$, $5.45$, $5.66$, $5.48$, and $5.09$, respectively; under the regular training state, the validation MAE values are $4.97$, $6.97$, $5.15$, $5.35$, $5.08$, and $5.03$, respectively. These results indicate that on Dataset II, the gap between Ours and the strongest purely data-driven baseline is small, but Ours still remains competitive with FRF, BiLSTM, Mamba, and Transformer while maintaining a high and stable $R^2$. In other words, the value of the proposed method on this dataset is better understood in terms of balanced performance and stable behavior rather than a large absolute numerical margin.

On the more challenging Dataset III, the supplementary results make the difference between methods more visible. As shown in Fig.~\ref{fig:datatrainadd}, the training/validation MAE of Ours are $4.42$ and $4.46$, respectively, lower than the $4.86$ and $5.38$ of FRF; the corresponding training/validation $R^2$ values are $0.76$ and $0.75$, higher than the $0.71$ and $0.64$ of FRF. Furthermore, from the integrated results in Fig.~\ref{fig:contrastnew}(C), under the further-optimized training state, the validation MAE values of CNN, BiLSTM, Mamba, Transformer, FRF, and Ours are $6.20$, $6.87$, $5.88$, $6.38$, $6.28$, and $4.67$, respectively; under the regular training state, they are $5.74$, $6.92$, $5.50$, $4.94$, $5.38$, and $4.46$, respectively. These results suggest that when the distribution gap between the validation and training sets becomes larger, the structural priors and constrained correction mechanism of the proposed model may provide more effective support.

A further comparison among Ours, FRF, and the four purely data-driven baselines shows that the relative value of Ours is not merely reflected in an occasional optimum under one training setting, but more in its relatively smooth performance changes across the two training states. Taking MAE as an example, when switching from the further-optimized training state to the regular training state, the validation MAE of Ours changes from $4.99 \rightarrow 4.69$, $5.09 \rightarrow 5.03$, and $4.67 \rightarrow 4.46$ on Dataset I, Dataset II, and Dataset III, respectively, with relatively small variation. In contrast, the corresponding changes of FRF are $5.51 \rightarrow 5.17$, $5.48 \rightarrow 5.08$, and $6.28 \rightarrow 5.38$, while some purely data-driven methods fluctuate even more across training states, for example, Transformer on Dataset III changes from $6.38$ to $4.94$, and BiLSTM on Dataset II changes from $8.07$ to $6.97$. This indicates that the final performance of purely data-driven models is more sensitive to the training state, whereas Ours shows relatively better stability under different training conditions.

Combined with the structure of the proposed model, the above supplementary results are consistent with the interpretation that an important part of the performance gain of PD-SOVNet comes from its structured modeling of the wheel--rail vibration transmission process. Specifically, the shared second-order vibration kernels explicitly introduce modal-response priors in the order domain; the MIMO coupling branch characterizes the basic transmission relationships among the four channels; the adaptive physical correction branch performs constrained compensation for sample-related modal shifts and coupling changes; and the time-domain branch further supplements the dynamic information not fully covered by explicit physical modeling. For this reason, the proposed method can achieve relatively stable validation performance while maintaining strong fitting ability on all three datasets, especially under more complex scenarios such as Dataset III. Overall, this supplementary experiment supports the previous main comparison results and helps reduce potential misinterpretation caused by incomplete figure presentation.

\subsubsection{Supplementary experiment: ablation analysis of the physical branch and time-domain Mamba}

To further clarify the actual role of each component in PD-SOVNet, we additionally compare three settings based on the complete model: retaining only the time-domain Mamba branch, retaining only the physical branch, and the complete model. Here, the ``physical branch'' includes all physical modules designed in this work, namely the shared second-order vibration kernels, cross-channel MIMO coupling, and the corresponding structured modeling components; the ``time-domain branch'' refers to the Mamba-based branch for supplementing temporal dynamics. The purpose of this supplementary experiment is not to construct new comparison methods, but to further analyze the relative contributions of frequency-domain physical modeling and time-domain dynamic modeling to the final performance.

\begin{table}[t]
    \centering
    \caption{Ablation results of the physical branch and time-domain Mamba}
    \label{tab:ablation_phys_mamba}
    \setlength{\tabcolsep}{5pt}
    \renewcommand{\arraystretch}{1.1}
    \resizebox{\linewidth}{!}{%
    \begin{tabular}{cccccccc}
        \toprule
        \multirow{2}{*}{Physical branch} & \multirow{2}{*}{Time-domain Mamba} 
        & \multicolumn{2}{c}{Dataset I}
        & \multicolumn{2}{c}{Dataset II}
        & \multicolumn{2}{c}{Dataset III} \\
        \cmidrule(lr){3-4} \cmidrule(lr){5-6} \cmidrule(lr){7-8}
        & & MAE $\downarrow$ & $R^2$ $\uparrow$
          & MAE $\downarrow$ & $R^2$ $\uparrow$
          & MAE $\downarrow$ & $R^2$ $\uparrow$ \\
        \midrule
        $\times$ & $\checkmark$ & 6.156 & 0.456 & 6.242 & 0.435 & 5.781 & 0.555 \\
        $\checkmark$ & $\times$ & 5.132 & 0.645 & 5.223 & 0.621 & 4.895 & 0.692 \\
        $\checkmark$ & $\checkmark$ & \textbf{4.992}& \textbf{0.663 }& \textbf{5.086} & \textbf{0.658} &\textbf{ 4.674} & \textbf{0.726} \\
        \bottomrule
    \end{tabular}%
    }
\end{table}
Table~\ref{tab:ablation_phys_mamba} reports the ablation results of the three settings on the three datasets. Overall, the complete model achieves the best results on all three datasets. More importantly, when the time-domain Mamba branch is removed, the performance decreases, but the degradation is relatively limited; in contrast, when the physical branch designed in this work is removed and only the time-domain Mamba branch is retained, the performance degrades much more significantly. These ablation results are consistent with the interpretation that the physics-driven branch provides the main structured modeling capacity in the current framework, while the Mamba temporal branch acts mainly as a supplementary component for residual dynamics.

On Dataset I, when only the time-domain Mamba branch is retained, the model achieves an MAE of $6.156$ and an $R^2$ of $0.456$; when only the physical branch is retained, the MAE decreases to $5.132$ and the $R^2$ increases to $0.645$; after further adding the time-domain Mamba branch, the complete model reduces the MAE to $4.992$ and improves the $R^2$ to $0.663$. This shows that on Dataset I, the physical branch itself already provides the main effective modeling capability, while the improvement brought by the Mamba branch is mainly reflected in further reducing the error and improving the fitting quality, with a much smaller contribution than that of the physical branch itself.

A similar trend can be observed on Dataset II. When only the time-domain Mamba branch is retained, the MAE is $6.242$ and the $R^2$ is only $0.435$; after retaining only the physical branch, the MAE decreases significantly to $5.223$ and the $R^2$ increases to $0.620$; the complete model further achieves an MAE of $5.086$ and an $R^2$ of $0.658$. Compared with Dataset I, the improvement of the complete model over the physical branch alone is slightly more obvious on Dataset II, especially in the further improvement of $R^2$. This indicates that on this dataset, the time-domain branch can supplement part of the dynamic information not fully covered by the physical branch, but the main performance still relies on the order-domain structural representation provided by the physical branch.

The same pattern still holds on the more complex Dataset III. When only the time-domain Mamba branch is retained, the model obtains an MAE of $5.781$ and an $R^2$ of $0.555$; when only the physical branch is retained, the MAE improves to $4.895$ and the $R^2$ increases to $0.692$; the complete model further achieves an MAE of $4.674$ and an $R^2$ of $0.726$. It can be seen that even in a more challenging scenario such as Dataset III, the physical branch still provides the main support for performance, while the time-domain Mamba branch further improves the accuracy and fitting ability on this basis. This also indicates that the proposed model does not simply rely on time-domain sequence modeling to achieve improvement, but instead obtains more robust prediction through the collaborative modeling of physical priors and temporal dynamic information.

When this ablation result is considered together with the previous main comparison experiments, the relative role of different components can be understood more clearly. First, the best results of the complete model on all three datasets indicate that relying only on a single time-domain modeling branch is insufficient to support the best performance, which is also consistent with the phenomenon that purely data-driven models are more easily affected by distribution changes under complex cross-wheel scenarios. Second, when only the physical branch is retained, the model already achieves results that are clearly better than those of the ``time-domain Mamba only'' setting, indicating that an important part of the gain is associated with explicit physical modeling, rather than simply with adding a stronger temporal network. Furthermore, the complete model still shows stable improvement over the physical-branch-only setting, which suggests that the role of the time-domain Mamba branch in this framework is mainly supplementary: it corrects and refines the temporal dynamics not fully described by the physical backbone, rather than dominating the overall prediction.

From the perspective of the method mechanism, this result is understandable, and is also similar to the experimental phenomenon reported in another study \cite{SUN2025112587}, where better performance is achieved only when the frequency domain and time domain are combined. In this work, the proposed physical branch explicitly introduces second-order vibration-response priors in the order domain and structurally models the transmission relationships among the four wheels through cross-channel MIMO coupling, so that it can more directly characterize the dominant mapping between wheel roughness values and axle-box vibration. In contrast, although the time-domain Mamba branch (its main role is further discussed in Section~\ref{section:noise}) can supplement the dynamic dependency information in the raw sequence, without the structural constraints provided by the physical backbone, its modeling process is still closer to general temporal fitting, and thus it is difficult for it alone to achieve optimal performance under cross-dataset and cross-wheel-group conditions. The reason why the complete model can maintain the best performance on all three datasets is precisely that the physical branch provides stable backbone modeling ability, while the time-domain Mamba branch supplements additional dynamic information on this basis, so that the two branches form a collaborative rather than substitutive relationship.

Overall, this ablation experiment further suggests that the structured representation capability of the physics-driven branch is an important source of performance, while the time-domain Mamba branch brings further but relatively moderate gains on top of it. Although this ablation does not isolate every design choice in the physical branch, it still provides supportive evidence that the improvement is not explained by the time-domain branch alone.

\subsubsection{Supplementary experiment: ablation of MIMO coupling and adaptive physical correction}
\begin{table}[t]
    \centering
    \caption{Ablation results of MIMO coupling and adaptive physical correction.}
    \label{tab:ablation_mimo_adaptive}
    \setlength{\tabcolsep}{5pt}
    \renewcommand{\arraystretch}{1.1}
    \resizebox{\linewidth}{!}{%
    \begin{tabular}{cccccccc}
        \toprule
        \multirow{2}{*}{MIMO coupling} & \multirow{2}{*}{Adaptive correction}
        & \multicolumn{2}{c}{Dataset I}
        & \multicolumn{2}{c}{Dataset II}
        & \multicolumn{2}{c}{Dataset III} \\
        \cmidrule(lr){3-4} \cmidrule(lr){5-6} \cmidrule(lr){7-8}
        & & MAE $\downarrow$ & $R^2$ $\uparrow$
          & MAE $\downarrow$ & $R^2$ $\uparrow$
          & MAE $\downarrow$ & $R^2$ $\uparrow$ \\
        \midrule
        $\times$ & $\times$ & 5.332 & 0.619 & 5.384 & 0.618 & 5.031 & 0.650 \\
        $\times$ & $\checkmark$ & 5.263 & 0.625 & 5.313 & 0.621 & 4.983 & 0.672 \\
        $\checkmark$ & $\times$ & 5.052 & 0.643 & 5.141 & 0.639 & 4.759 & 0.689 \\
        $\checkmark$ & $\checkmark$ & \textbf{4.992} & \textbf{0.663} & \textbf{5.086} & \textbf{0.658} & \textbf{4.674} & \textbf{0.726} \\
        \bottomrule
    \end{tabular}%
    }
\end{table}
To further clarify the respective roles of the cross-channel MIMO coupling and the adaptive physical correction branch, we additionally conduct a $2\times2$ ablation study by removing each module separately and jointly. Specifically, four variants are compared: (1) removing both MIMO coupling and adaptive correction, (2) retaining only adaptive correction, (3) retaining only MIMO coupling, and (4) using the complete model. The corresponding results are summarized in Table~\ref{tab:ablation_mimo_adaptive}.

Overall, the complete model achieves the best results on all three datasets, indicating that the combination of MIMO coupling and adaptive correction provides the strongest performance. More importantly, the ablation trends show that the two modules do not contribute equally. Among them, MIMO coupling provides the larger portion of the gain, while the adaptive correction branch brings a further but more moderate improvement. This observation is consistent with the design of PD-SOVNet: the structured physical backbone is mainly established by the shared vibration-response priors together with the cross-channel coupling mechanism, whereas the adaptive correction branch mainly serves as a constrained sample-dependent refinement around that backbone.

On Dataset I, removing both modules leads to an MAE of 5.332 and an $R^2$ of 0.619. Retaining only adaptive correction slightly improves the results to 5.263 and 0.625, whereas retaining only MIMO coupling yields a more noticeable improvement to 5.052 and 0.643. After further combining both modules, the full model reaches the best performance, with an MAE of 4.992 and an $R^2$ of 0.663. This indicates that, on Dataset I, the main gain already comes from introducing cross-channel structured coupling, while the adaptive correction provides an additional refinement on top of it.

A similar trend can be observed on Dataset II. Starting from the variant without either module (MAE 5.384, $R^2$ 0.618), adding only adaptive correction results in a limited improvement (MAE 5.313, $R^2$ 0.621), while adding only MIMO coupling produces a larger gain (MAE 5.141, $R^2$ 0.639). The complete model further improves the results to an MAE of 5.086 and an $R^2$ of 0.658. Therefore, on Dataset II, the adaptive correction still provides a positive contribution, but its role remains supplementary relative to that of the MIMO coupling module.

The same pattern remains valid on the more challenging Dataset III. Without either module, the model achieves an MAE of 5.031 and an $R^2$ of 0.650. Retaining only adaptive correction reduces the MAE slightly to 4.983 and raises the $R^2$ to 0.672, while retaining only MIMO coupling leads to a more substantial improvement, with an MAE of 4.759 and an $R^2$ of 0.689. After combining both modules, the full model reaches the best overall performance, with an MAE of 4.674 and an $R^2$ of 0.726. Compared with the first two datasets, the additional gain brought by adaptive correction is more visible on Dataset III, suggesting that sample-dependent structured refinement becomes more helpful when the validation distribution is more challenging and modal shifts or coupling variations are harder to capture using the shared physical backbone alone.

Taken together, these results support two conclusions. First, the cross-channel MIMO coupling is the more influential component among the two modules and constitutes an important part of the structured modeling capacity of PD-SOVNet. Second, although the gain brought by adaptive correction is moderate rather than dominant, it is consistently positive across all three datasets and becomes more visible on the more challenging dataset. This is also consistent with its intended role in the model design: adaptive correction is not introduced to replace the physical backbone with a strong free residual branch, but to provide controlled sample-dependent compensation for local modal shifts and coupling variations. Therefore, the best performance is achieved when both modules are retained, indicating that structured coupling and constrained adaptive refinement are complementary rather than substitutive.
\subsection{Noise Injection Experiments on the Training Set}\label{section:noise}

\begin{figure}
    \centering
    \includegraphics[width=1.\linewidth]{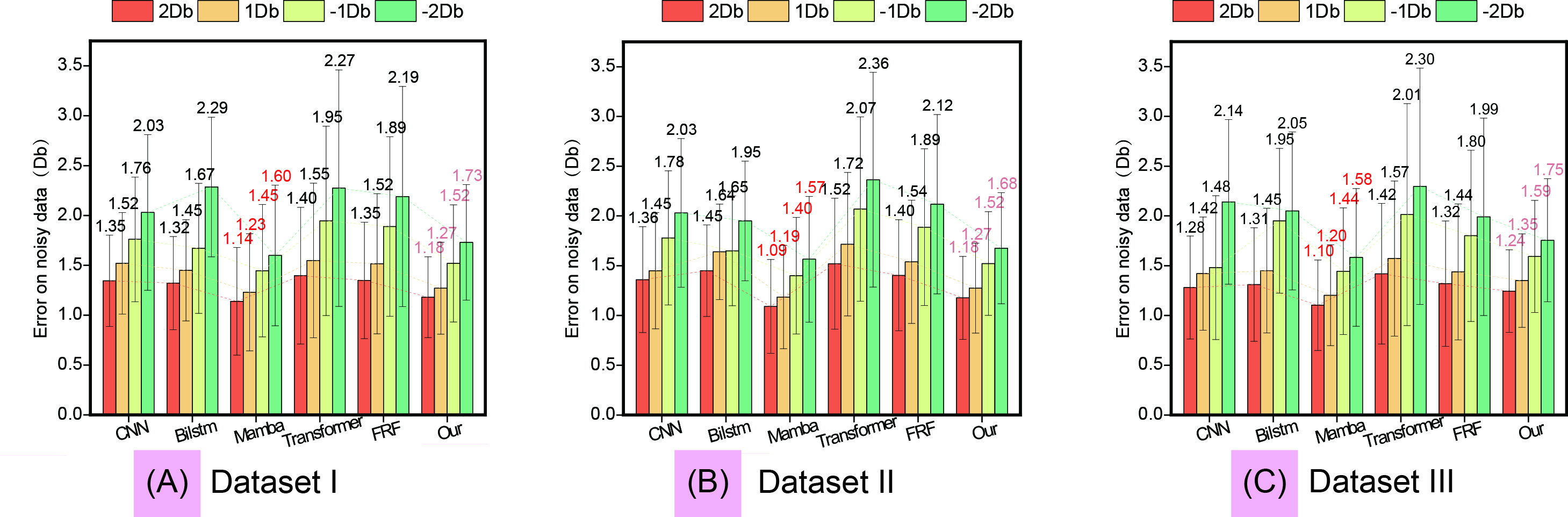}
    \caption{Comparison results of the training-set noise injection experiments. (A)--(C) correspond to Dataset I, Dataset II, and Dataset III, respectively. Each subfigure shows the error variations of the six methods under four noise conditions ($2$\,dB, $1$\,dB, $-1$\,dB, and $-2$\,dB), where the bar chart denotes the mean error and the error bars indicate the fluctuation range. All models are evaluated using the weights saved when the training MAE first decreased to approximately $0.8$, so as to support an aligned comparison under a common training reference. This figure is used to compare the stability and noise robustness of different methods under perturbed inputs.}
    \label{fig:noiseresult}
\end{figure}

Considering that real operational vibration signals already contain measurement noise and operating-condition disturbances, while the current training data still do not cover all railway lines or all possible noise patterns, relying only on the natural noise contained in the training set is insufficient to fully reflect the stability of the models when facing unseen perturbations during deployment. On the other hand, existing studies have shown that models with explicit physical constraints or physically parameterized structures often suffer from relatively obvious performance degradation when facing additional noise and model mismatch \cite{noise_1,noise_2}. Based on this observation and the requirements of deployment, this work further conducts controlled noise injection experiments on the training set to evaluate the error variation trends of different models under additional perturbations. For all models, the weights saved when the training \textbf{MAE first dropped to 0.8} are selected as a common reference state, and on this basis the input samples are perturbed with four noise levels, namely $2$ dB, $1$ dB, $-1$ dB, and $-2$ dB. MAE is used as the evaluation metric to compare CNN, BiLSTM, Mamba, Transformer, FRF, and the proposed method (Ours) on Dataset I, Dataset II, and Dataset III, as shown in Fig.~\ref{fig:noiseresult}.

From the overall trend, as the noise intensity increases, the MAE of all models on the three datasets shows an increasing trend, indicating that additional noise continuously degrades the stability of roughness regression. The noise-injection results suggest a trade-off. In this specific perturbation test, pure Mamba shows the strongest absolute noise robustness, whereas PD-SOVNet provides a better balance between physically guided structure and robustness, consistently improving over the FRF baseline while remaining competitive with the stronger data-driven baselines. Therefore, this experiment should be interpreted as evidence that the Mamba branch helps compensate for the noise sensitivity of the physical architecture, rather than as evidence that the proposed method is uniformly optimal under all perturbation conditions.

\begin{figure}
    \centering
    \includegraphics[width=1\linewidth]{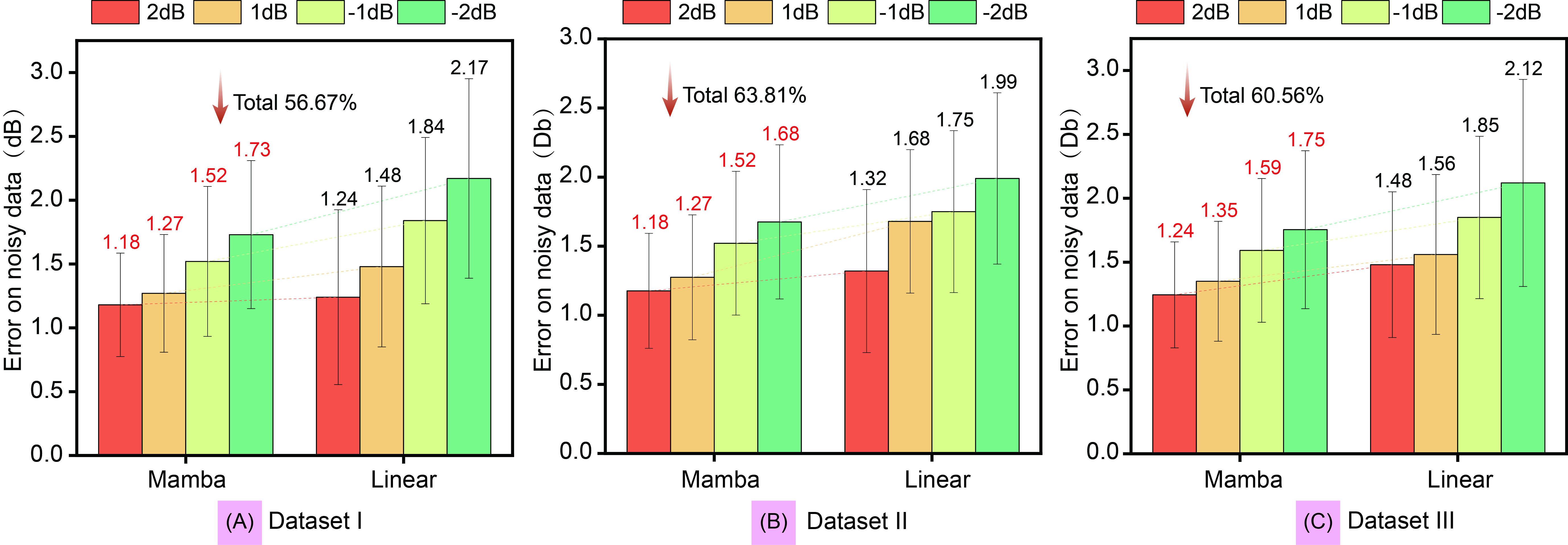}
    \caption{Comparison of the noise ablation results of the temporal branch. (A)--(C) show the MAE performance of the proposed model on Dataset I, Dataset II, and Dataset III, respectively, when the temporal branch is implemented using Linear and Mamba under four levels of input noise perturbation ($2$ dB, $1$ dB, $-1$ dB, and $-2$ dB). The bar chart denotes the mean error under different noise conditions, the error bars represent the corresponding fluctuation range, and the numerical labels indicate the specific MAE values under each noise level. The dashed lines are used to assist in visualizing the trend of error variation as the noise increases. It can be seen that under all three datasets and all four noise conditions, the model error with the Mamba temporal branch is consistently lower than that with the Linear branch, indicating that stronger temporal modeling capability can effectively alleviate the sensitivity of the physics-driven architecture to additional input perturbations near the deployment weights, thereby improving overall stability.}
    \label{fig:noise_ablation}
\end{figure}

From the perspective of individual datasets, on \textbf{Dataset I}, all models show increasing error as the noise grows, but the growth rates of Mamba and Ours are relatively slower. Specifically, under the four noise levels, the MAE values of Mamba are $1.14$, $1.23$, $1.45$, and $1.60$, while those of Ours are $1.18$, $1.27$, $1.52$, and $1.73$, both of which are clearly better than CNN ($1.35$, $1.52$, $1.76$, $2.03$), BiLSTM ($1.32$, $1.45$, $1.67$, $2.29$), Transformer ($1.40$, $1.55$, $1.95$, $2.27$), and FRF ($1.35$, $1.52$, $1.89$, $2.19$). On \textbf{Dataset II}, this trend becomes even more obvious. The MAE values of Mamba under the four noise levels are $1.09$, $1.19$, $1.40$, and $1.57$, while those of Ours are $1.18$, $1.27$, $1.52$, and $1.68$; in contrast, FRF increases to $1.40$, $1.54$, $1.89$, and $2.12$, and Transformer reaches $1.52$, $1.72$, $2.07$, and $2.36$. On \textbf{Dataset III}, although the absolute error levels of all models differ slightly from those on the first two datasets, the overall ranking remains basically unchanged: Mamba achieves $1.10$, $1.20$, $1.44$, and $1.58$, while Ours achieves $1.24$, $1.35$, $1.59$, and $1.75$, still clearly outperforming CNN, BiLSTM, Transformer, and FRF. Across the three datasets, although the proposed method does not surpass pure Mamba, it is the most stable among the methods containing physical priors and consistently remains in a strong second tier on all three datasets.

A further comparison among Ours, FRF, and the four purely data-driven baselines shows that Ours has better noise robustness than the purely physics-based baseline FRF. For example, on Dataset I, the MAE values of FRF under the four noise levels are $1.35$, $1.52$, $1.89$, and $2.19$, whereas the corresponding values of Ours are $1.18$, $1.27$, $1.52$, and $1.73$; on Dataset II, FRF yields $1.40$, $1.54$, $1.89$, and $2.12$, while Ours reduces them to $1.18$, $1.27$, $1.52$, and $1.68$; on Dataset III, FRF gives $1.32$, $1.44$, $1.80$, and $1.99$, whereas Ours achieves $1.24$, $1.35$, $1.59$, and $1.75$. This indicates that when modeling relies only on physical structure, the model is more likely to be affected by additional perturbations under noisy conditions. In contrast, by introducing a temporal compensation branch, the proposed method improves its ability to absorb noise-related residuals while preserving physical constraints. On the other hand, compared with purely data-driven baselines such as CNN, BiLSTM, and Transformer, Ours maintains lower error across all three datasets and all four noise levels, indicating that physical priors still play a positive role in regression stability under perturbed inputs. However, in terms of absolute values, pure Mamba remains the most robust model in this experiment, which also suggests that the noise sensitivity of the current physics-driven structure has not yet been completely eliminated.

To further analyze the influence of the temporal branch on noise robustness, Fig.~\ref{fig:noise_ablation} presents the results obtained by replacing the temporal branch in the proposed model from Linear to Mamba, which also indirectly supports the phenomenon discussed in \cite{noise_1,noise_2}. It can be observed that under the four noise levels on all three datasets, the use of Mamba consistently reduces the error to different degrees. Taking Dataset I as an example, under the four noise levels, the MAE decreases from $1.24$, $1.48$, $1.84$, and $2.17$ for the Linear branch to $1.18$, $1.27$, $1.52$, and $1.73$ for the Mamba branch, with a cumulative relative reduction of $56.67\%$. On Dataset II, the MAE decreases from $1.32$, $1.68$, $1.75$, and $1.99$ to $1.18$, $1.27$, $1.52$, and $1.68$, with a cumulative reduction of $63.81\%$; on Dataset III, the MAE decreases from $1.48$, $1.56$, $1.85$, and $2.12$ to $1.24$, $1.35$, $1.59$, and $1.75$, with a cumulative reduction of $60.56\%$. Averaged over the three datasets, the cumulative reduction under the four noise levels is about $60\%$, corresponding to an average relative improvement of about $15\%$ per noise level. These results indicate that after replacing the simple linear temporal branch with Mamba, the sensitivity of the model to input perturbations is alleviated in a relatively stable manner.

Combined with the structure of the proposed method, the above phenomenon can be understood from two aspects. First, the physics-driven second-order vibration operator branch can explicitly model modal responses, transmission relationships, and cross-channel coupling, and therefore has strong structural constraint capability under noise conditions similar to those in the training set. However, when additional unseen perturbations appear in the input, explicit physical modeling cannot fully cover all noise-related dynamics, and the resulting error is easily amplified. Second, the Mamba temporal branch can further capture more complex temporal dependencies and residual dynamics from the raw angle-domain sequence, thereby compensating for the noise-related components that are not sufficiently characterized by the physical branch. Overall, the results of this section do not imply that the physics-driven structure has completely solved the noise sensitivity issue, but they do suggest that strengthening the temporal branch can alleviate part of the performance degradation of the physical architecture under noisy conditions.
\section{Discussion}

This study should be interpreted as an application-oriented investigation of physics-guided roughness regression rather than as a complete physical reconstruction of the wheel--rail system. The main takeaway is that embedding simplified modal-response priors into a gray-box architecture can help stabilize prediction behavior under the current unseen-wheel protocol, especially when the validation distribution becomes more challenging.

At the same time, the present results do not justify claims of universal superiority across broader operating conditions. The scope of the current conclusions is limited by at least three aspects: the physical branch remains a low-dimensional approximation of the vibration process; the available data cover a relatively narrow speed range and a unified vehicle architecture; and the input--label pairing is constrained by the practical difficulty of synchronizing operational vibration records with wheel-profile measurements. Therefore, the reported results should be understood as evidence of feasibility under the current data protocol rather than as final deployment-level validation.

From an industrial perspective, the value of the present study lies not only in the specific network structure, but also in showing that wheel-group-based evaluation, physically guided modeling, and data-closed-loop thinking can be combined within a practical monitoring workflow. At the same time, a more fundamental issue reflected by the experiments is that, including the proposed method, existing models still find it difficult to stably maintain a very low MAE under cross-wheel scenarios. This suggests that the current bottleneck may not lie solely in the network architecture itself, but is also likely constrained by the coverage of the training data, the reliability of the labels, and the quality of the input--label pairing. More specifically, if the training set does not sufficiently cover response distributions under different wheel groups, different operating conditions, and different roughness-evolution states, then even if the model architecture is further deepened or complicated, its generalization ability may not continue to improve.

This observation further points to a potentially more practically meaningful future direction, namely constructing a large-scale data closed loop for engineering scenarios. A related practical difficulty is that vibration data can be continuously collected during train operation relatively easily, but the corresponding wheel roughness labels are difficult to obtain at high frequency. Even if wheel-profile data can be collected through an underfloor lathe or wayside inspection system, it is still not easy to achieve strict temporal alignment and wheel-group matching with operational vibration samples. This is related not only to limitations in acquisition conditions and maintenance cycles, but also to the complexity of cross-department collaboration and data-management processes. Therefore, in engineering practice, the real difficulty is not merely ``collecting more vibration data,'' but rather ``continuously collecting and correctly matching high-quality input--label pairs.''

The discussion of TSAD is therefore included not as a direct extension of the main regression claim, but as an application-oriented perspective on how future data acquisition, relabeling, and sample prioritization may be improved in real maintenance systems. One feasible direction is to introduce a time-series anomaly detection (TSAD) model to actively screen samples during operation that are obviously inconsistent with the existing training distribution, and mark them as potential ``high-value supplementary samples.'' The significance of such a mechanism is not to directly replace the roughness regression model, but to assist in constructing a more efficient data-acquisition closed loop. On the one hand, it can help identify new operating conditions, new wheel--rail states, or abnormal response patterns that the current model has not sufficiently covered. On the other hand, anomaly scores or similar confidence outputs can also provide a basis for evaluating model outputs, as well as for subsequent manual review and sample-priority ranking, thereby improving the efficiency of additional data collection and label matching.

\begin{figure}[H]
    \centering
    \includegraphics[width=1\textwidth]{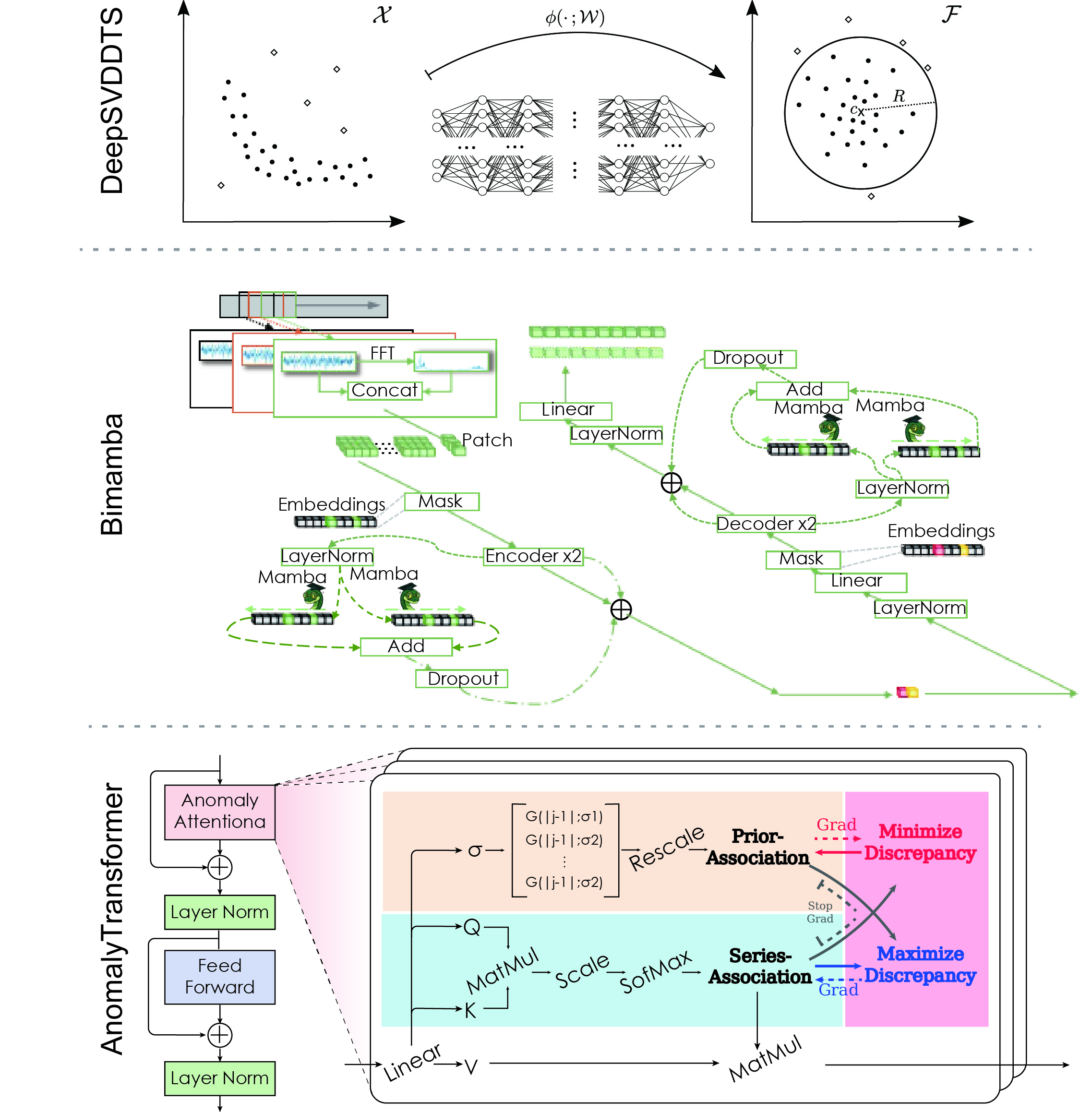}
    \caption{Illustration of time-series anomaly detection models for future active data acquisition and abnormal-sample screening. The figure shows three candidate models, including DeepSVDDTS, BiMamba, and Anomaly Transformer. The purpose of introducing this figure in the Discussion section is not to rank their structures, but to illustrate that anomaly detection can be used as an auxiliary module in the data closed loop to identify high-value samples that are inconsistent with the current training distribution, and to provide a reference for subsequent supplementary sampling and label matching.}
    \label{fig:anomaly_future}
\end{figure}

As shown in Fig.~\ref{fig:anomaly_future}, in the future, multiple time-series anomaly detection models may be considered as auxiliary tools in the data acquisition process. The key point here is not which specific detection architecture is used, but rather making use of the ``out-of-distribution sample identification'' capability provided by anomaly detection to actively discover, from an engineering perspective, those data segments that are most difficult for the current regression model to handle but are also the most valuable. From the methodological perspective, this idea is also consistent with the results of this work: the current upper bound of model performance is constrained not only by the network structure, but also by the coverage of the training distribution. Therefore, instead of simply continuing to increase model complexity, constructing a closed-loop pipeline of ``anomaly detection screening -- high-value sample recovery -- label matching -- regression model updating'' may be a more engineeringly feasible future path.

Overall, under the current data conditions, the present results suggest the feasibility of physics-guided gray-box modeling for wheel polygon roughness regression, but its room for improvement is still jointly constrained by the precision of physical modeling and the coverage of the training data. Future work needs, on the one hand, to further strengthen the explicit modeling of the wheel--rail coupling process, and, on the other hand, to establish a more stable mechanism for data acquisition and updating in practical engineering.

\begin{figure}[H]
    \centering
    \includegraphics[width=1\linewidth]{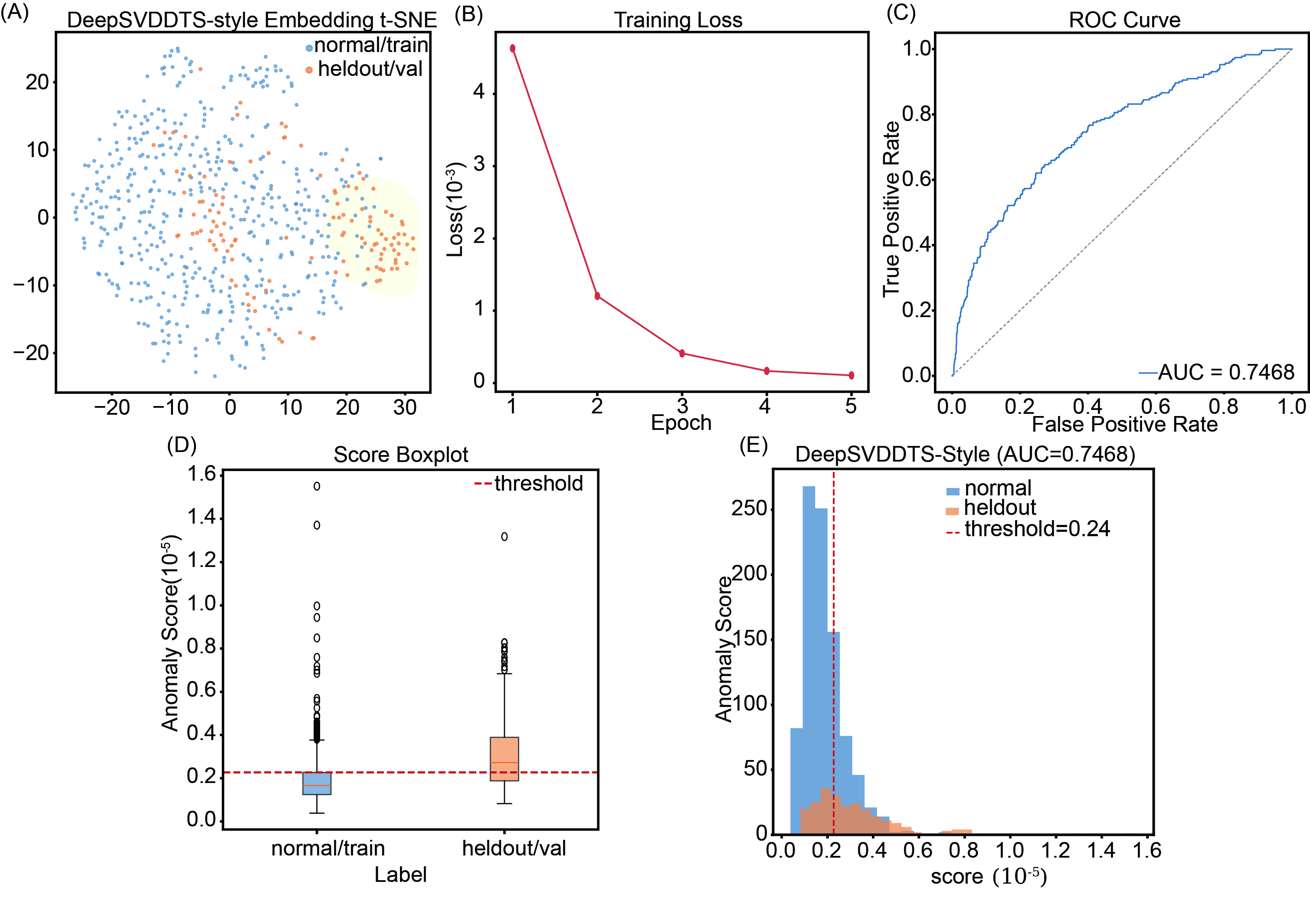}
    \caption{Visualization analysis of DeepSVDDTS-style anomaly detection results on training samples and held-out samples. (A) shows the sample distribution in the embedding space for observing the overall shift between the training distribution and the held-out samples. (B) shows the convergence process of the training loss. (C) gives the ROC curve and the corresponding AUC result. (D) shows the boxplot of anomaly scores for the two types of samples. (E) gives the prediction confusion matrix based on threshold partitioning. (F) shows the anomaly-score distribution and the corresponding threshold position. This figure is used to illustrate that, for the ``training distribution consistency detection'' task considered in this work, the anomaly scores already exhibit certain statistical differences between the training samples and the held-out samples, thus providing a feasible screening signal for subsequent active data supplementation.}
    \label{fig:tsad_deepsvdd}
\end{figure}

In addition to the model structure itself, the experiments in this work also indicate that the current performance upper bound of the task is, to a large extent, jointly constrained by the coverage of the training data and the difficulty of input--label matching. Therefore, a direction that is more meaningful from an engineering perspective than simply continuing to deepen the regression model is to construct an active screening mechanism for data supplementation, that is, to continuously collect large amounts of vibration data during operation and prioritize those data segments that are inconsistent with the current training distribution and therefore are more worthy of additional labeling. In this sense, the TSAD discussed in this work is not equivalent to the traditional TSAD task of ``detecting fault points in normal sequences,'' but is closer to a form of \emph{training distribution consistency detection}. Its goal is to determine whether a new sample deviates from the response distribution covered by the existing training set, and further provide support for subsequent data recovery, manual review, and label matching. The three models are trained using the angle-domain-resampled training set as normal training samples and the angle-domain-resampled validation set as anomaly-detection samples.

First, from the DeepSVDDTS-style results shown in Fig.~\ref{fig:tsad_deepsvdd}, this type of method shows certain feasibility for the current task. The embedding distributions, anomaly-score boxplots, and score histograms all indicate that there is a certain degree of distribution difference between the training samples and the held-out samples. Although this difference is not completely separable, it already forms a usable shift in a statistical sense. In particular, the mean, variance, and tail range of the score distribution show certain differences between the two types of samples. This implies that the anomaly score can be used not only as a ranking basis for individual samples, but also as a statistical quantity for the overall state change of data over a period of time, thereby improving the diagnostic value of ``whether additional data are needed.'' Furthermore, the ROC curve and confusion matrix also indicate that, although such methods have not achieved very strong discriminative ability, they can already provide a kind of coarse-grained, low-cost prior signal for active screening. For the data-supplementation scenario emphasized in this work, such a capability itself is meaningful, because its goal is not to precisely locate faults, but to discover sample regions that are ``inconsistent with the existing training set'' as much as possible.

\begin{figure}[H]
    \centering
    \includegraphics[width=1\linewidth]{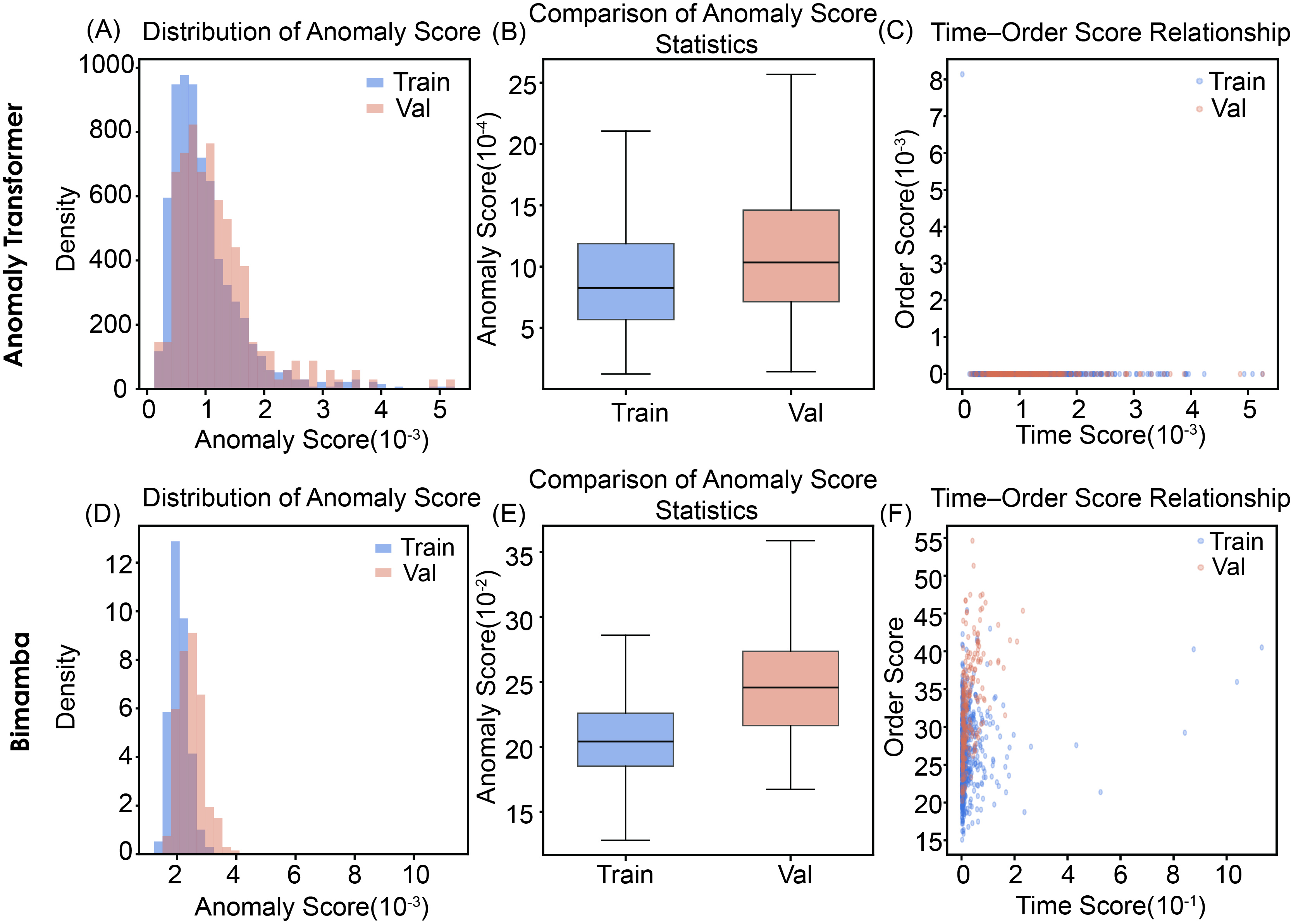}
    \caption{Comparison results of anomaly scores between training samples and validation samples for Anomaly Transformer and BiMamba. (A)--(C) correspond to Anomaly Transformer, including the anomaly-score distribution, statistical boxplots of the anomaly scores, and the relationship between time-domain scores and order-domain scores; (D)--(F) correspond to BiMamba, including the same three types of visualizations. This figure is used to compare the suitability of different time-series anomaly detection models for the task of ``training distribution consistency detection,'' and to illustrate that the differences in the mean, variance, and joint distribution of anomaly scores between training samples and validation samples can serve as reference signals for subsequent active sample screening and data supplementation.}
    \label{fig:tsad_compare}
\end{figure}

Second, Fig.~\ref{fig:tsad_compare} presents comparative results of Anomaly Transformer and BiMamba on this task. From Fig.~\ref{fig:tsad_compare}(A)--(C), it can be seen that the anomaly-score distributions of Anomaly Transformer still have considerable overlap between the training and validation samples, the distribution shift in the boxplots is relatively limited, and the relationship between the time-domain scores and the order-domain scores does not exhibit a particularly clear structural pattern. This indicates that its suitability for the current task is relatively limited. A possible reason is that Anomaly Transformer is more suitable for detection scenarios in which local anomalies coexist with normal backgrounds within long sequences, whereas the scenario considered in this work is not to locate anomalous segments within a single long sequence, but to determine whether an entire sample deviates from the current training distribution. In other words, the two tasks are not fully consistent at the semantic level, and therefore directly transferring its anomaly-modeling mechanism may lead to limited effectiveness.

In contrast, the BiMamba results shown in Fig.~\ref{fig:tsad_compare}(D)--(F) exhibit more obvious score-shift phenomena. Whether in the anomaly-score distributions, the medians and dispersions in the boxplots, or the joint distribution of the time-domain scores and the order-domain scores, the validation samples show a clearer overall shift relative to the training samples. This indicates that BiMamba can at least more stably perceive whether the current sample still lies close to the training distribution. Similar to the DeepSVDDTS-style results, the key point here is also not to identify some specific fault point by point, but to use the changes in the mean, variance, and high-score regions of the anomaly scores to make a preliminary judgment as to whether a sample belongs to ``high-value supplementary data.'' From this perspective, BiMamba appears to be relatively more suitable for the scenario considered in this work.

Based on the above observations, a cautious but positive judgment can be made regarding the data-supplementation idea proposed in this work: from the perspective of feasibility, it is highly likely to be valid to use TSAD-like methods to assist in identifying data segments that are inconsistent with the training distribution. The fundamental reason is that the difficulty of the current cross-wheel roughness regression task does not come entirely from insufficient model expressiveness, but also from insufficient coverage of the real operational distribution by the training samples. Therefore, as long as the anomaly detection model can statistically distinguish between ``samples commonly seen in the training set'' and ``samples that clearly deviate from the training distribution,'' it can provide value for active engineering data acquisition. Furthermore, if statistical quantities such as the mean, variance, and tail proportion of the anomaly scores, together with measures such as KL divergence and expert heuristic rules, are jointly used, it may be possible to construct a more interpretable sample-priority evaluation mechanism, thereby improving the efficiency of subsequent label acquisition and cross-department data matching.

Of course, this idea still has its boundaries. First, the above results mainly indicate that \emph{distribution differences can be detected to some extent}, but are still insufficient to show that anomaly scores can strictly correspond to a specific physical operating-condition change or a specific source of roughness variation. Second, the current experiments are still based on the existing data protocol and have not yet formed a truly online closed-loop data-supplementation pipeline, so their engineering value still needs to be further verified in longer-term and more realistic data-collection processes. Finally, different anomaly detection models do not show consistent suitability for this task, which also indicates that ``directly reusing existing TSAD methods'' may not be sufficient. A more reasonable future direction may be to design a dedicated detection framework for out-of-training-distribution sample screening that simultaneously incorporates time-domain, order-domain, and physical-prior information.

Overall, this subsection suggests that, in the wheel polygon roughness regression task, time-series anomaly detection can be understood not only as an auxiliary analysis tool, but also as an important component for constructing a data closed loop. Its value does not lie in replacing the main regression model, but in helping the engineering system actively discover the data samples that are most worthwhile to supplement, and in improving efficiency, especially in cross-department collaboration scenarios. If this mechanism can be integrated with subsequent sample matching, manual review, and model-updating processes, it may help alleviate the current difficulty of obtaining large-scale, high-quality input--label pairs, and further improve the practical applicability of cross-wheel roughness regression models, thereby laying a solid foundation for future real-world deployment.

\section*{Declaration of competing interest}
The authors declare that they have no known competing financial interests or personal relationships that could have appeared to influence the work reported in this paper.
\section*{Declaration of Generative AI and AI-assisted technologies in the manuscript preparation process}
During the preparation of this work, the authors used ChatGPT (OpenAI) in order to improve language readability and polish expression. After using this tool, the authors reviewed and edited the content as needed and take full responsibility for the content of the manuscript.

\section*{Acknowledgments}
This work is supported by the National Key Research and Development Program of China (2023YFB3308100), the China State Railway Group Co., Ltd. Science and Technology Research and Development Program Project (K2024J011) and the Natural Science Foundation of Shandong Province (ZR2023ME124).

\section{Conclusion}

This paper presented PD-SOVNet, a physics-guided gray-box framework for estimating 1st--40th-order wheel roughness spectra from axle-box vibrations. Under the current real-world data protocol and unseen-wheel evaluation setting, the method achieved competitive accuracy and relatively stable cross-wheel performance, with its clearest advantage observed on the more challenging dataset. The results suggest that combining structured physical priors with learned temporal dynamics is a promising direction for industrial roughness regression. At the same time, the current study remains limited by the available speed range, dataset coverage, and the practical difficulty of input--label matching. Future work will therefore focus on broader-condition validation, stronger module-level verification, and more scalable data-closed-loop mechanisms for engineering deployment.



\bibliographystyle{elsarticle-num} 
\bibliography{reference}






\end{document}